\pdfoutput=1
\documentclass{article}

\usepackage[preprint, nonatbib]{neurips_2022}
\usepackage[numbers]{natbib}

\usepackage[utf8]{inputenc} 
\usepackage[T1]{fontenc}    
\usepackage{hyperref}       
\usepackage{booktabs}       
\usepackage{amsfonts}       
\usepackage{nicefrac}       
\usepackage{microtype}      
\usepackage{xcolor}         
\usepackage{xurl}
\usepackage{microtype}
\usepackage{graphicx}
\usepackage{booktabs} 
\usepackage{svg}
\usepackage{amsmath}
\usepackage{pifont}
\usepackage{rotating}
\usepackage{tikz,pgfplots} 
\pgfplotsset{compat=1.12}
\pgfplotsset{plot coordinates/math parser=false}
\usepackage{subcaption}
\usepackage{multicol}
\usepackage{stfloats}
\usepackage{caption}
\usetikzlibrary{patterns}
\usepackage{collectbox}

\newcommand{\mybox}{%
    \collectbox{%
        \setlength{\fboxsep}{1pt}%
        \fbox{\BOXCONTENT}%
    }%
}
\newcommand\newsubcap[1]{\phantomcaption%
       \caption*{\figurename~\thefigure(\thesubfigure): #1}}
\newcommand{\textcircledcenter}[1]{\raisebox{.5pt}{\textcircled{\raisebox{-.9pt} {#1}}}}
\newcommand{\maxthru}{16\% }
\newcommand{\maxbatchsize}{2$\times$ }
\newcommand{\gptthru}{19\% }
\newcommand{\robertathru}{26\% }

\newcommand{\cmark}{\ding{51}}%
\newcommand{\xmark}{\ding{55}}%

\title{Tempo: Accelerating Transformer-Based Model Training through Memory Footprint Reduction}

\author{%
  Muralidhar Andoorveedu\textsuperscript{1}, Zhanda Zhu\textsuperscript{2,3}, Bojian Zheng\textsuperscript{1,3}, Gennady Pekhimenko\textsuperscript{1,3} \\
  \textsuperscript{1}University of Toronto, Toronto, Canada \\
  \textsuperscript{2}Shanghai Jiao Tong University, Shanghai, China \\
  \textsuperscript{3}Vector Institute, Toronto, Canada \\
  \texttt{\{andoorve, zhanda, bojian, pekhimenko\}@cs.toronto.edu} \\
}

\begin{document}

\maketitle

\begin{abstract}

Training deep learning models can be computationally expensive. 
Prior works have shown that increasing the batch size can potentially lead to better overall throughput. However, the batch size is frequently limited by the accelerator memory capacity due to the activations/feature maps stored for the training backward pass, as larger batch sizes require larger feature maps to be stored. Transformer-based models, which have recently seen a surge in popularity due to their good performance and applicability to a variety of tasks, have a similar problem. To remedy this issue, we propose \textit{Tempo}, a new approach to efficiently use accelerator (e.g., GPU) memory resources for training Transformer-based models. Our approach provides drop-in replacements for the GELU, LayerNorm, and Attention layers, reducing the memory usage and ultimately leading to more efficient training. 
We implement \textit{Tempo} and evaluate the throughput, memory usage, and accuracy/loss on the BERT$_{LARGE}$ pre-training task. We demonstrate that \textit{Tempo} enables up to \maxbatchsize higher batch sizes and \maxthru higher training throughput over the state-of-the-art baseline. We also evaluate \textit{Tempo} on GPT2 and RoBERTa models, showing \gptthru and \robertathru speedup over the baseline.
\end{abstract}
\section{Introduction}
\label{sec:Introduction}

Transformer-based models such as BERT \cite{devlin2018} and GPT-2 \cite{gpt2} have found success in numerous general natural language processing tasks including question answering~\cite{squad}, paraphrasing~\cite{mrpc}, natural language inference~\cite{mnli}, and even areas outside language tasks such as image recognition~\cite{image_recog}. However, training such models can be highly expensive in terms of time, monetary resources and carbon footprint \cite{datamovement, energyandpolicy}. For instance, the pre-training of BERT$_{LARGE}$ takes 4 days to complete on 16 Cloud TPUs (64 TPU chips total) \cite{devlin2018}, which costs about \$10,000 \cite{Sharir2020TheCO}. Training a more recent Transformer-based model, GPT-3, has an even more astonishing price tag - \$12 million\cite{gpt3cost}. Hence, even a small decrease in the end-to-end training time of Transformer-based models matters.

Although there has been significant progress made in accelerating Transformers using specialized hardware (e.g., Google TPUs~\cite{tpu}, NVIDIA Tensor Cores~\cite{v100}) in the past few years, a fundamental issue with Transformer-based models is that they are limited by the memory capacity of hardware accelerators. For example, even a batch size of 1 does not fit into a modern GPU with 12GB of memory when training BERT with sequence length 512~\cite{oom_issues}. Reducing memory footprint \cite{capuchin, checkpoint, vdnn} is a viable option to allow larger batch training, leading to better hardware utilization and ultimately improved training throughput \cite{zheng2020}. 

Many existing approaches to memory footprint reduction (e.g., offloading~\cite{vdnn, superneurons, capuchin}, checkpointing~\cite{checkpoint, zheng2020, dtr, checkmate}, and data compression/encoding~\cite{jain2018, actnn}) either have high computational overhead or do not apply to Transformer-based models directly. Prior approaches fall into two main categories, neither of which are satisfactory for the Transformer-based model case. First, these techniques may be too general~\cite{capuchin, actnn, zeroinf, dtr, checkmate} to utilize the specifics of Transformer-based models well, such as the multi-headed attention mechanism used in Transformers~\cite{vaswani2017}, or optimization opportunities available in specific layers such as the LayerNorm~\cite{layernorm} layer. For example, although checkpointing~\cite{checkpoint, checkmate} can significantly enlarge batch size, it also brings high overhead (e.g., 30\% performance degradation observed in some prior works \cite{checkpoint}).
Second, if prior works are specific, they focus on other types of models/layers with ideas not being applicable to Transformers. For example, Gist and In-Place ABN deal with CNNs \cite{jain2018, rotabulo2017}. 

In our work, we demonstrate that low overhead memory footprint reduction can lead to a positive improvement in throughput. In addition, unlike prior works which do not leverage the specifics of Transformer-based models, we propose a new approach specifically tailored for Transformer-based models, called \textit{Tempo}. This approach includes three new techniques: (i) In-place GELU, (ii) In-place LayerNorm, and (iii) Sub-Layer Dropout Recomputation. In-place GELU and In-place LayerNorm both use alternative derivations for the computation of the backward passes of these layers. These derivations allow some activations that are normally retained during the forward pass (to be later used in the backward pass) to be discarded, leading to a more memory-efficient implementation. Sub-Layer Dropout Recomputation discards activations within the high memory footprint attention mechanism during the forward pass, then recomputes these during the backward pass without recomputing extra unnecessary tensors. \textit{Tempo} is able to increase the training throughput with larger batch sizes by reducing the total memory footprint of the models during training. To our best knowledge, this is the first work to explore memory footprint optimizations specifically for Transformer-based layers that show not just footprint reduction, but the actual increase in throughput using the extra memory savings. 
Tempo reduces the memory footprint of training Transformer-based models by targeting a major part of the total footprint -- the \textit{activation memory} \cite{tbd} (the saved feature maps during the forward pass of the model that are required for backpropagation \cite{backprop}).
All the proposed techniques provide a large memory footprint reduction with very low throughput degradation (as low as 1\%). Our results show up to \maxbatchsize improvement in batch size for BERT$_{LARGE}$ pre-training at a sequence length of 512 on modern GPUs while increasing the training throughput by up to \maxthru.

\section{Background and Motivation}
\label{sec:Background_and_Motivation}

\subsection{Memory Footprint of BERT}

BERT \cite{devlin2018} is a popular natural language processing model that is based on the Transformer architecture \cite{vaswani2017}. The model has been successfully applied to a variety of tasks such as question answering (SQuAD \cite{squad}), paraphrasing (MRPC~\cite{mrpc}), natural language inference (MNLI~\cite{mnli}), and others~\cite{Socher2013RecursiveDM, Zellers2018SWAGAL} through a two step training process. The training process entails first training on a general unlabelled data set (\emph{pre-training}) \cite{devlin2018}. The faster second part of the training process (\emph{fine-tuning}) takes the parameter weights produced by the pre-training section and further trains on a downstream task such as question answering \cite{squad} or sentiment analysis \cite{Socher2013RecursiveDM} which it accomplishes through the addition of a specialized output layer \cite{devlin2018}. 

The BERT architecture allows for multiple different configurations depending on model hyperparameters selected, some being derived from the original Transformers paper; these include the hidden layer size ($H$), sequence length ($S$), number of attention heads ($A$) and number of layers ($L$).

\begin{figure}[!t]
    \centering
    \includegraphics[width=0.60\columnwidth]{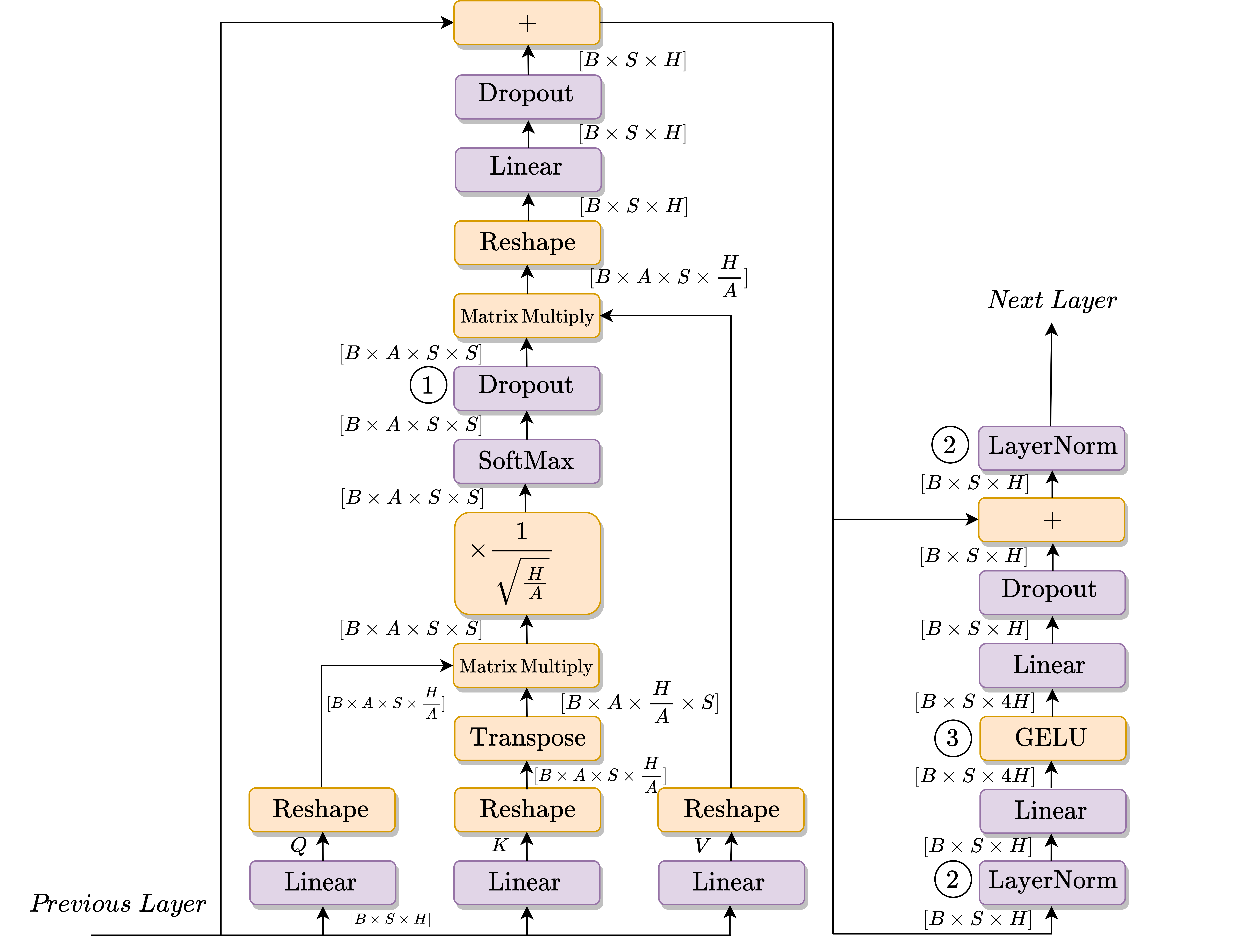}
    \caption{A diagram of a single Transformer encoder \cite{vaswani2017} layer used in BERT \cite{devlin2018}. This is based on the Huggingface implementation of BERT \cite{huggingface}. As in the BERT paper, $A$ represents the number of attention heads, and $H$ represents the hidden size. We represent the batch size by $B$ and the sequence length by $S$. Sizes of intermediate tensors (both retained activations and unretained intermediates) are annotated.}
    \label{fig:bert_diagram}
\end{figure}

In the context of this work, we point out some of the relevant parts of the model and their activation memory footprint with respect to these hyperparameters referring to Figure~\ref{fig:bert_diagram}. 

\textcircledcenter{{1}} At this point, where attention \cite{vaswani2017} is calculated we observe that the size of each of the feature maps goes as $\mathcal{O}(S^2) - $ there are a variety of previous techniques and models that have been explored in the literature to deal with this problem \cite{survey}. Additionally, at this point note that we store three feature maps of size $[B \times A \times S^2]$. Calculations based on Figure \ref{fig:bert_diagram} at the BERT$_{BASE}$ parameters show that at a sequence length of 512 these three feature maps \textbf{account for $56\%$ of the encoder layer activation memory.}

\textcircledcenter{{2}}
At these two points, we store the input to the two LayerNorm layers of size $[B \times S \times H]$

\textcircledcenter{{3}}
Here a GELU \cite{hendrycks2016} layer is used as the activation function for the preceding fully-connected layer of size $[B \times S \times 4H]$. The activation memory for this function stores \textbf{almost $17\%$ of the total layer activation memory of} BERT$_{BASE}$ at a sequence length of 128.

\subsection{Why Activation Memory Matters}

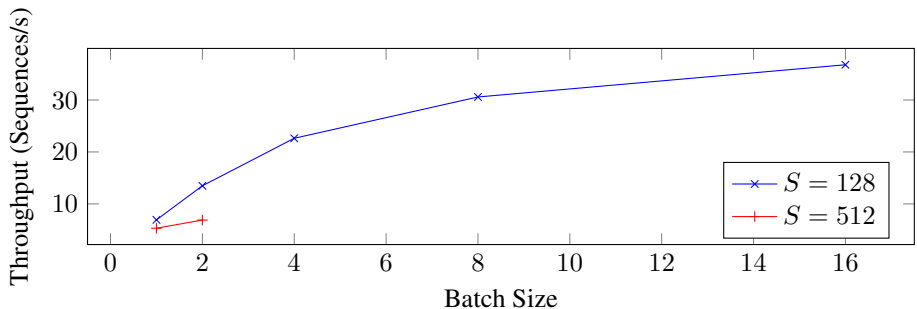
\begin{figure}[!htbp]  
\centering 
    \begin{tikzpicture}
    \begin{axis}[
            width=0.9\columnwidth,
            height=0.3\textwidth,
            xlabel=Batch Size,
            ylabel=Throughput (Sequences/s),
            legend pos = south east,
        ]
        \addplot[color=blue, mark=x] coordinates {(1, 6.93) (2, 13.48) (4, 22.63) (8, 30.58) (16, 36.78)};
        \addplot[color=red, mark=+] coordinates {(1, 5.32) (2, 6.89)};
        \legend{$S=128$, $S=512$}
    \end{axis}
\end{tikzpicture}
\caption{Plots of throughput (sequences/s) vs batch size for BERT$_{LARGE}$ \cite{devlin2018} fine-tuning on the MRPC \cite{mrpc} task at sequence lengths 128 and 512 on four 2080Ti \cite{2080ti} GPUs. The maximum batch sizes are respectively 16 and 2.}
\label{fig:throughput}
\end{figure}

As iterated in previous works \cite{vdnn, jain2018, capuchin, zheng2020, actnn} there are multiple benefits to reducing the memory footprint of models. 
First, it allows for larger models which can positively affect the model's performance on downstream tasks \cite{devlin2018}. 
Second, memory footprint reduction can allow for a larger batch size. This, in turn, could lead to better utilization of the GPU hardware \cite{batchsize}, increasing the overall throughput \cite{zheng2020}. In order to verify this possibility for Transformer-based models, we conduct our own experiments using Huggingface's BERT implementation \cite{huggingface} to train BERT${_{LARGE}}$ on the MRPC \cite{mrpc} fine-tuning task. Figure~\ref{fig:throughput} shows the throughput on this task for sequence lengths of 128 and 512. From the figure, we conclude that there is a steady improvement in batch size when the sequence length is 128. This is also the case when the sequence length is 512, however, in this situation the trend ends more abruptly as the memory consumption of the model exceeds the GPU memory capacity, showing a clear opportunity to take advantage of memory footprint reduction.



We note that previous works on Transformer-based models show that although the model parameters contribute to the memory footprint, the main memory capacity consumer during training is actually the activation feature maps \cite{tbd, checkmate, checkpoint, capuchin, jain2018, dtr, actnn}. In addition, the majority of this activation memory will be used in each of the BERT Transformer encoder layers. Profiling the Huggingface BERT$_{BASE}$ implementation \cite{huggingface} on the MRPC \cite{mrpc} fine-tuning task at a batch size of 32 and sequence length of 128 shows that 66\% of the total memory is taken up by these encoder activations. More details on this are shown in Appendix \ref{sec:memory}.

\subsection{Key Prior Works}

There are three major prior techniques used in training memory footprint reduction of deep learning models. The first of these is \emph{Checkpointing} \cite{checkpoint, dtr, checkmate, zheng2020}. This technique involves discarding certain feature maps in the forward pass while retaining others. Later, in the backward pass, these discarded feature maps may be recomputed from the retained feature maps, and thus used in the computation of the gradients. The second technique is \emph{Offloading} \cite{capuchin, vdnn, superneurons}. In this case, the main idea involves taking feature maps that would be stored in the GPU memory, and instead offloading them to the CPU memory. These techniques can also involve \emph{pre-fetching} tensors from the CPU memory in anticipation of their use. Offloading suffers from a dependence on system variables such as the communication channel bandwidth \cite{vdnn, capuchin}. It also requires extensive engineering effort to avoid high overhead \cite{actnn}. Finally, \emph{Compression/encoding}; this can be divided into two different categories, lossless and lossy \cite{jain2018, actnn}. However, the fundamental idea is to compress, or reduce the space taken up by feature maps in the forward pass, then decompress it for use in the backward pass. 


These techniques are usually largely orthogonal to one another as was shown in prior works where both offloading and checkpointing are used simultaneously~\cite{capuchin, superneurons}. We expand on these techniques in Appendix \ref{sec:methods}.

\subsection{Why Tempo?}

Although the techniques in the previous section show good performance on a variety of models, they suffer from a variety of issues. Checkpointing's scope is often too broad to consider certain layer-specific optimizations and alternative derivations that can provide lower overhead \cite{rotabulo2017}. Furthermore, overhead can be high (as much as 30\%) \cite{checkpoint}. Offloading can be system- dependent and requires significant engineering effort, while compression can be lossy or not applicable to the Transformer case.
Hence, there is a clear need for a deeper look at activation memory optimizations for Transformer-based neural networks in particular. To our best knowledge, our work is the first to explore such optimizations tuned to improving the throughput of Transformer-based models. Table~\ref{tab:comparison_table} shows a summary comparison of Tempo and various other techniques, with the major points that differentiates our technique from prior work.

\begin{table}[!htbp]
    \centering
    \tiny
\begin{tabular}{rccccc}
\toprule
\\[0.25in]
Feature &
\begin{rotate}{60} Capuchin \end{rotate} &
\begin{rotate}{60} Checkmate \end{rotate} &
\begin{rotate}{60} ActNN  \end{rotate} &
\begin{rotate}{60} Gist \end{rotate} &
\begin{rotate}{60} \textbf{Tempo} \end{rotate} \\ \midrule

Layer-Specific                & \xmark & \xmark & \xmark & \cmark & \cmark \\ 
Transformer-Specific          & \xmark & \xmark & \xmark & \xmark & \cmark \\
Lossless                      & \cmark & \cmark & \xmark & $\sim$\footnotemark & $\sim$\footnotemark  \\
Drop-In Layer Replacement     & \xmark & \xmark & \cmark & \cmark & \cmark \\ 
Online                        & \cmark & \xmark & \cmark & \cmark & \cmark \\ \bottomrule
\end{tabular}
    \caption{Comparison between Tempo and Capuchin \cite{capuchin}, Checkmate \cite{checkmate}, ActNN \cite{actnn}, and Gist~\cite{jain2018}.}
    \label{tab:comparison_table}
\end{table}
\footnotetext[1]{Some of the Gist \cite{jain2018} optimizations are lossy.}
\footnotetext[2]{Accuracy of our lossy optimization is tunable, offering a flexible tradeoff between the accuracy and the hardware cost.}

\vspace{-1em}
\section{Tempo: Key Ideas}
\label{sec:Key_Ideas}

We now present the major ideas that lays behind the design of Tempo: (1) \textbf{In-place GELU}, (2) \textbf{In-place LayerNorm}, and (3) \textbf{Sub-Layer Dropout Recomputation}. The major theme behind all of these ideas is to compute the backward pass as normal, while \emph{using less storage} to do so. To this end, In-place GELU and In-place LayerNorm compute the output of each layer \emph{in-place}; instead using the output activation to compute the gradient. Sub-Layer Dropout Recomputation also discards the output, and through a closer look at the structure of the Dropout layer is able to recompute the output without excessive recomputation. We strongly suggest reading Appendix \ref{appdix:FirstImplDetails} for the implementation details. We also add in this appendix a new optimization of softmax that we use that further reduces memory \cite{deberta_huggingface}.

\subsection{In-place GELU} 
\begin{figure}[!htbp]
\begin{subfigure}{0.45\columnwidth}
    \centering
    \includegraphics[width=\columnwidth]{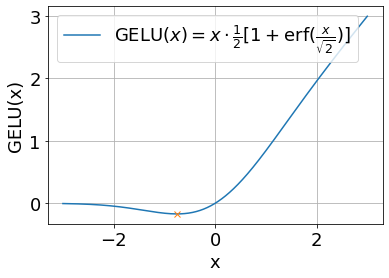}
    \newsubcap{A plot of the GELU\cite{hendrycks2016} function near the origin, along with the marked minimum point.}
    \label{fig:gelu_plot}
\end{subfigure}
\begin{subfigure}{0.45\columnwidth}
    \centering
    \includegraphics[width=\columnwidth]{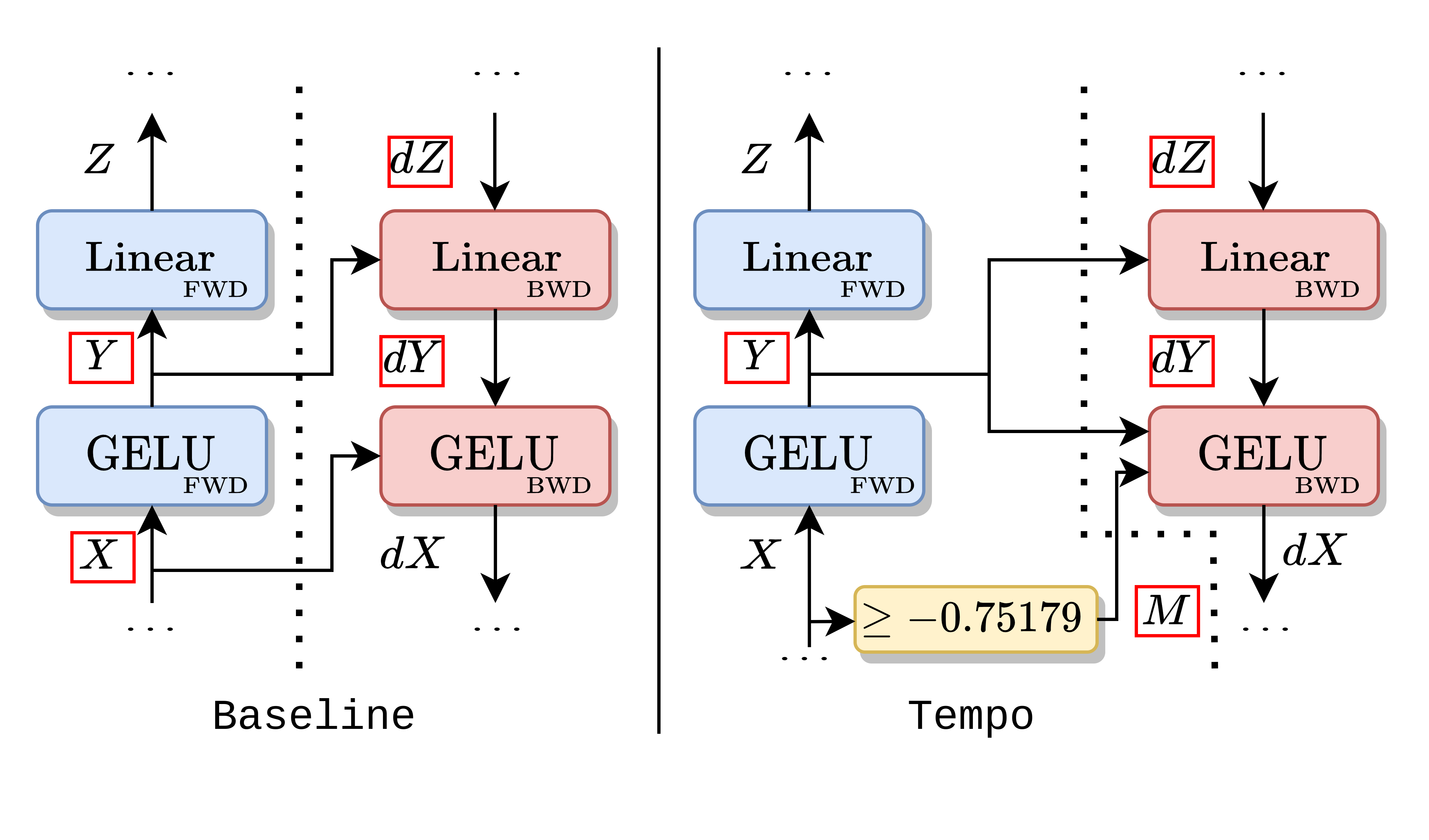}
    \newsubcap{Saved feature maps between the baseline and Tempo. Note that our method only saves a 8-bit mask\protect\footnotemark[3] that denotes whether the input is greater or less than the minimum value, instead of the full 32-bit input feature map.}
    \label{fig:gelu_compare}
\end{subfigure}
\end{figure}

The GELU layer is used as an activation function for the feed-forward section of the BERT layer (\textcircledcenter{3} in Figure~\ref{fig:bert_diagram}) \cite{devlin2018}. A plot of this function is shown in Figure~\ref{fig:gelu_plot}. Referring to the baseline in Figure~\ref{fig:gelu_compare}, note that both $X$ and $Y$ are stored for the backward pass. $Y$ is needed for the downstream fully connected layer, while $X$ is stored for the GELU layer itself \cite{pytorch}. Prior work has demonstrated that certain activation functions such as ReLU may be computed in-place \cite{jain2018}. This can be done without affecting the calculation of the backward pass. If we were able to compute the GELU function in-place, potentially by recovering the input from the output on the backward pass, we could save the storage required for $X$.
However, this is impossible to do directly. A key observation to make with respect to the GELU function is that it is not bijective -- hence there is no function that will be able to compute the input from the output without additional information.


However, we observe that the GELU function is both continuous and has only one extremum, a minimum value at $x \approx -0.75179$ as can be seen in Figure~\ref{fig:gelu_plot}. Notably, this implies that just one additional piece of information: which \textbf{side of the minimum} the input originates from, allows us to compute the inverse of the GELU. This is because on each side of the minimum the function is one-to-one, and hence the input is recoverable from the output in \emph{each section}. Based on this key observation, we can discard the input, and simply retain the output of the GELU, as well as the additional information on whether the input is greater than or equal to the value at which the minimum occurs. Figure~\ref{fig:gelu_compare} illustrates the difference between our method and the baseline. 

In order to execute this efficiently on a real system, we note that the original derivative in terms of the input can be composed with the function inverse in order to create a composite kernel. This kernel consists of a polynomial approximation of this composite function, the approximation being necessary since GELU is transcendental, and therefore the inverse cannot be solved in terms of elementary functions \cite{transcendental}. Further details are discussed in Appendix \ref{appdix:FirstImplDetails}.

\footnotetext[3]{Pytorch boolean masks use 8-bits per value \cite{pytorch}. Masks can also be implemented as 1-bit manually but this brings extra overhead due to unpacking and packing bit tensors.}

\subsection{In-place LayerNorm}

The LayerNorm layer is used at multiple points in the Transformer encoder layer \cite{vaswani2017}, which we denote by \textcircledcenter{2} in Figure~\ref{fig:bert_diagram}. Usually, the gradient computation of LayerNorm relies on the gradient input from the next layer, as well as the input feature map which is stashed for this computation \cite{pytorch}.

Similar to GELU, we are able to derive an expression for the gradient of the LayerNorm layer \emph{as a function of its output}. In this context, the output of LayerNorm must be stashed to compute the gradient of the successive fully connected layer anyways.  Using this approach, the memory footprint overhead of LayerNorm is just the intermediate mean and variance computed in the forward pass. The full derivation is presented in Appendix \ref{appdix:FirstImplDetails} which is extended from the treatment of BatchNorm in \cite{rotabulo2017}.


\textbf{Comparison with Checkpointing}: Note that although In-place GELU requires more memory compared to recomputing $Y$ from $X$, it will have increased overhead due to the recomputation. Additionally, our technique is orthogonal to conventional Checkpointing, as it could take advantage of the fact that no recomputation is required for the input $X$ for both In-place GELU and In-place LayerNorm.

\subsection{Sub-Layer Dropout Recomputation}
In this section we explore the idea of sub-layer granularity checkpointing, or partial recomputation applied to the Dropout layer \cite{dropout} found in \textcircledcenter{1} in Figure~\ref{fig:bert_diagram}. The function of a dropout layer is to set the output of $p\%$ of entries in the incoming feature map to zero (``drop'' the outputs) and then scale the remaining outputs by the factor $\frac{1}{1-p}$, which makes the network less sensitive to any output of the preceding feature map, thereby making it more robust~\cite{dropout}. 

\begin{figure}[!htbp]
    \centering
    \includegraphics[width=0.75\columnwidth]{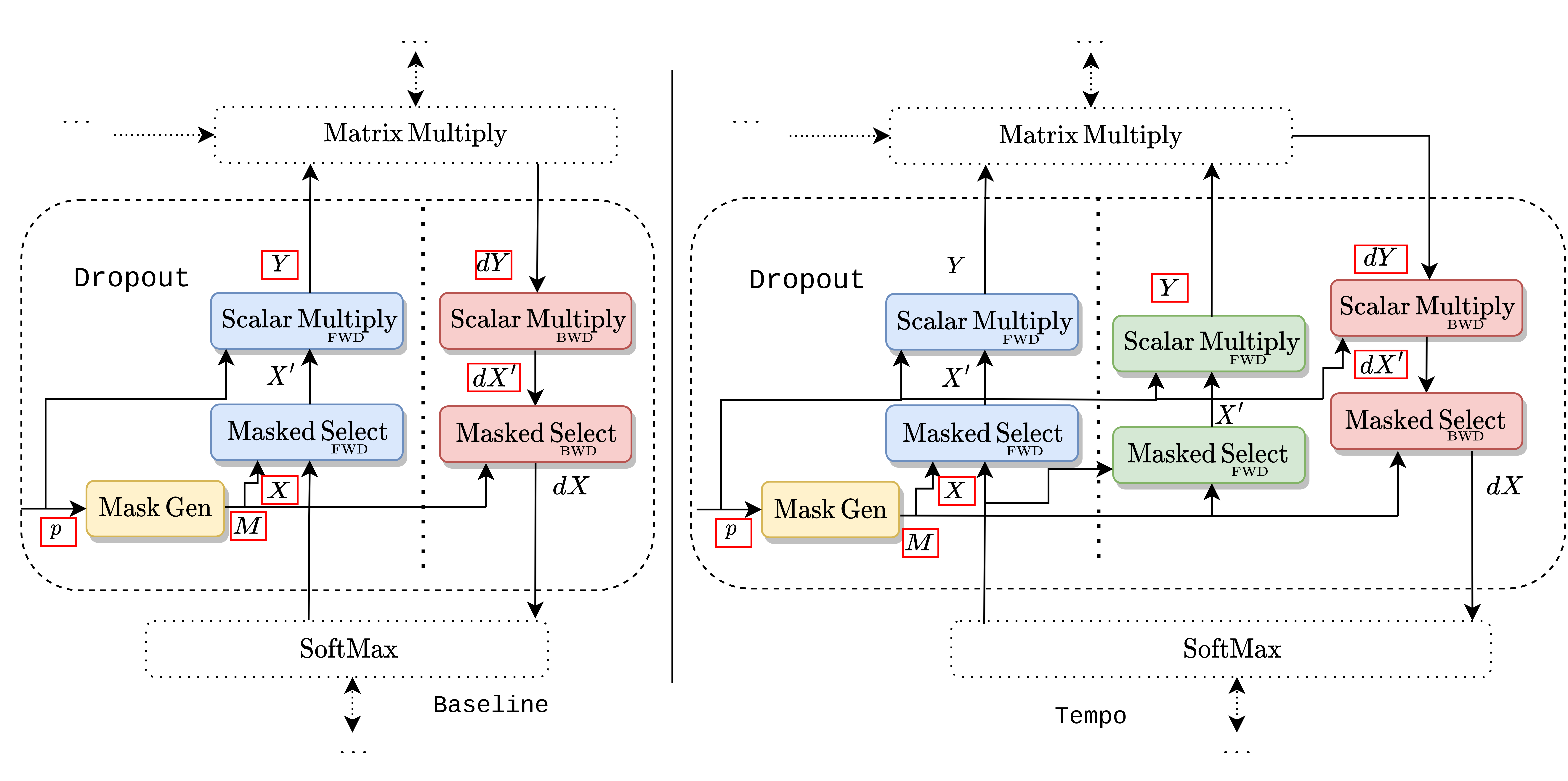}
    \caption{Comparison of dropout implementation between the baseline, and our method. Note that we only save the mask, and recompute the other output. The representation on the left is not an exact copy of the PyTorch implementation, rather it is an illustrative representation.}
    \label{fig:dropout}
\end{figure}

We define sub-layer recomputation as a technique where recomputation of only \emph{some} of the feature maps is necessary for the backward pass that may be produced by a given layer's output. We observe that better recomputation strategies are possible if we carefully deconstruct layers in the case where they store multiple outputs. This observation can be directly applied to the Dropout layer. In the computation of Dropout, both a mask (which records the entries which are set to zero in implementations of Dropout \cite{pytorch, mxnet}) and output are produced. If a layer-based checkpointing implementation \cite{check_pytorch} was used, it would cause both the mask and the output to be recomputed in the backward pass if the layer is checkpointed, thus requiring higher overhead. However, we notice that nothing precludes us from simply doing only one of these recomputations. Storing the mask would only reduce the recomputation (including memory transfer) time, while the fact that the mask itself only has Boolean values allows us to keep most of the memory benefit of recomputation. In this way, we can save the storage required for the output at the critical $\mathcal{O}(S^2)$ Attention section (\textcircledcenter{3} in Figure~\ref{fig:bert_diagram}) for the cost of a simple mask multiply. This technique is illustrated in Figure \ref{fig:dropout}.

\subsection{Other Engineering Optimizations}
We note that PyTorch uses a memory-inefficient implementation of the softmax function which retains both the input and output of the function for the backward pass \cite{pytorch}. Instead, only the output is necessary. This optimization has also previously been implemented as part of some models in the Huggingface library \cite{deberta_huggingface}. We use this optimization as well in our implementation of the attention mechanism to further reduce the activation memory pressure.

\vspace{-1em}
\section{Evaluation}
\label{sec:Evaluations}

\pgfplotsset{every axis/.append style={
                    axis line style={<->}, 
                    xlabel={Sequence Length},          
                     legend style={font=\small}
                    }}

\subsection{Methodology}

\paragraph*{Infrastructure}
    Our main test setup consists of 4 NVIDIA RTX 2080 Ti GPUs \cite{2080ti}, each with 11 GB of memory connected over PCIe v3 \cite{pcie}. We also use an Amazon Web Services p3.8xlarge \cite{aws_p3} instance consisting of 4 NVIDIA Tesla V100 GPUs \cite{v100} each with 16 GB of memory connected using NVLink \cite{nvlink}. For our ablation studies, we employ a system with an NVIDIA A100 GPU \cite{a100} with 40 GB of memory. We summarize the detailed setup in Appendix \ref{appdix:Setup}.
    

\paragraph*{Applications}
We evaluate our work using both the BERT pre-training and fine-tuning tasks~\cite{devlin2018}. For pre-training, we employ the NVIDIA DeepLearningExamples library~\cite{deeplearningexamples} with the English Wikipedia dataset \cite{wikipedia}. We perform the training in two phases, the first (i.e., longer) phase at a sequence length of 128, and the second (i.e., shorter) phase at a sequence length of 512 \cite{devlin2018, deeplearningexamples}. For throughput and memory experiments, we use the BERT$_{LARGE}$ configuration. For our fine-tuning task, we use the MRPC \cite{mrpc} paraphrasing task on BERT$_{BASE}$ using the Huggingface library \cite{huggingface}.

For our ablation studies, we also train both RoBERTa \cite{liu2019roberta} and GPT2\cite{gpt2}. For the evaluation of RoBERTa, we use the Fairseq library \cite{ott2019fairseq}, while GPT2 uses the Huggingface GPT2 model \cite{huggingface}. Both of these models use the WikiText Dataset for evaluation \cite{wikitext}.


\paragraph*{Metrics}

The first metric we focus on is the total \emph{memory footprint} of our method compared to the baselines. There are two ways to look at this metric.
First, we compare the maximum batch size possible for each method. We compare this across sequence lengths of 128 and 512 \footnote[5]{These are the sequence lengths of Phase 1 and Phase 2 of pre-training \cite{devlin2018, deeplearningexamples}.} on BERT$_{LARGE}$ for both 2080Ti and V100 GPUs. Second, we compare the total memory used by PyTorch at a given commonly used batch size for the same parameters. The second metric we use is the \emph{throughput} for which we count the total number of sequences per second processed.
Finally, we provide a comparison between our method and the baseline method on BERT$_{BASE}$ pre-training in order to compare the loss curves and show the change due to our lossy optimizations. We also provide fine-tuning curves on the MRPC \cite{mrpc} task, training for 10 epochs to ensure no significant accuracy deviations.

Our ablation studies only use the throughput metric.

\subsection{Results}

We use two major baselines. The first baseline is the NVIDIA BERT$_{LARGE}$ model~\cite{deeplearningexamples}, with no memory footprint techniques applied which we refer to as the \emph{Baseline}. The second one is the same model, with the default checkpointing applied, based on the PyTorch implementation, applied at the input of each Transformer encoder layer \cite{pytorch, deeplearningexamples} and is similar to the Huggingface implementation \cite{huggingface}. We refer to this baseline as \emph{Checkpoint}. We refer to our method that uses In-Place GELU, In-Place LayerNorm, Sub-Layer Dropout Recomputation, and the softmax engineering optimization as \emph{Tempo}.
    
    

\paragraph*{Impact on Memory Footprint}

Table~\ref{tab:batch_results} shows the maximum batch size and memory consumed at a fixed batch size for all three methods. Additionally, the total memory used at a batch size of 15 at a sequence length of 128 is 11.3 GB, 8.3 GB and 9.2 GB respectively for \emph{Baseline}, \emph{Checkpoint}, and \emph{Tempo}. From this, we conclude that \emph{Checkpoint} reduces the memory footprint to a much higher degree than both \emph{Baseline} and \emph{Tempo}. This is expected, as \emph{Checkpoint} discards most of the feature maps to be recomputed \cite{huggingface, deeplearningexamples} no matter the performance cost. \emph{Tempo} still provides a significant increase in batch size over \emph{Baseline} at the sequence length of 512 -- \textbf{we see 2$\times$ and 1.5$\times$ larger batches over \emph{Baseline}} for the 2080 Ti and V100 respectively but, as the next section shows, with much better throughput.

      \begin{table}[!htbp]
          \centering
         \scriptsize
          \begin{tabular}{cccc}
              \toprule
               Technique & Sequence Length & Batch Size \\
               \midrule
               Baseline & 128 & 15\\
               Baseline & 512 & 1\\
               Checkpoint & 128 & 50\\
               Checkpoint & 512 & 4\\
               Tempo & 128 & 24\\
               Tempo & 512 & 2\\
               \bottomrule
            \end{tabular}
            \begin{tabular}{cccc}
               \toprule
               Technique & Sequence Length & Batch Size \\
               \midrule
               Baseline & 128 & 28\\
               Baseline & 512 & 4\\
               Checkpoint & 128 & 96\\
               Checkpoint & 512 & 18\\
               Tempo & 128 & 41\\
               Tempo & 512 & 7\\
               \bottomrule
          \end{tabular}
          \caption{The maximum batch size on both 2080 Ti (left) and V100 (right) for BERT$_{LARGE}$.}
          \label{tab:batch_results}
      \end{table}

\pgfplotstableread[row sep=\\,col sep=&]{
    interval & Baseline & Checkpoint & Tempo    \\
    128 & 15 & 50 & 24 \\
    512 & 1 & 4 & 2 \\
    }\batchtwentyeighty

\pgfplotstableread[row sep=\\,col sep=&]{
    interval & Baseline & Checkpoint & Tempo    \\
    128 & 28 & 96 & 41 \\
    512 & 4 & 18 & 7 \\
    }\batchvhundred


\paragraph*{Impact on Throughput}

Figure~\ref{fig:thru_results} illustrates our main results with respect to throughput. From the figure, we can see that \emph{Tempo} outperforms both \emph{Checkpoint} and \emph{Baseline} across \textbf{both sequence lengths} and \textbf{across different hardware setups}. We observe an improvement of \textbf{16\%} over \emph{Baseline} on the 2080 Ti at a sequence length of 512. At these settings, we also have an improvement of 8\% over \emph{Checkpoint}. We also observe up to \textbf{27\%} over \emph{Checkpoint} on the V100 at a sequence length of 512, which also corresponds to a 5\% improvement over \emph{Baseline}. This is despite the fact that \emph{Checkpoint} uses the largest batch size as per Table~\ref{tab:batch_results}. This is because \emph{Checkpoint} stores feature maps at the beginning of each Transformer encoder layer, and recomputes these layers \cite{huggingface, deeplearningexamples}. Hence, an increased batch size also means more recomputation. In contrast, \emph{Tempo} is able to decrease the total memory footprint, and then convert this decrease into a substantial performance improvement over the \emph{Baseline} due to the use of only low overhead mechanisms.

\pgfplotstableread[row sep=\\,col sep=&]{
    interval & Baseline & Checkpoint & Tempo    \\
    128 & 142.3160342 & 120.0995009 & 147.74820384344287 \\
    512 & 23.37308406 & 25.10084511 & 27.21911522 \\
    }\thrutwentyeighty

\pgfplotstableread[row sep=\\,col sep=&]{
    interval & Baseline & Checkpoint & Tempo    \\
    128 & 171.2244967 & 139.1819184 & 176.2587747 \\
    512 & 36.04625143 & 29.83960227 & 37.82857622\\
    }\thruvhundred

\begin{figure}[!htbp]  
\centering 
  \begin{subfigure}[b]{0.495\columnwidth}
    \begin{tikzpicture}
    \begin{axis}[
            ybar,
            bar width=.2cm,
            width=1.05\columnwidth,
            height=0.65\columnwidth,
            symbolic x coords={128, 512},
            enlarge x limits=0.75,
            xtick=data,
            ymin=0,ymax=225,
            ylabel={ Throughput (Sequences/s)},
            cycle list={pattern=horizontal lines, , pattern=crosshatch},
        ]
        \addplot table[x=interval,y=Baseline]{\thrutwentyeighty};
        \addplot table[x=interval,y=Checkpoint]{\thrutwentyeighty};
        \addplot table[x=interval,y=Tempo]{\thrutwentyeighty};
        \node (1.04x) at (axis cs:128,180) {\scriptsize \mybox{1.04$\times$}};
        \node (1.08x) at (axis cs:512,60) {\scriptsize \mybox{1.08$\times$}};
        \legend{Base., Chk., Tempo}
    \end{axis}
\end{tikzpicture}
    \newsubcap{2080 Ti} \label{fig:M1}  
   \end{subfigure}
\begin{subfigure}[b]{0.495\columnwidth}
    \begin{tikzpicture}
    \begin{axis}[
            ybar,
            bar width=.2cm,
            width=1.05\columnwidth,
            height=0.65\columnwidth,
            symbolic x coords={128, 512},
            enlarge x limits=0.75,
            xtick=data,
            yticklabels={,,},
            ymin=0,ymax=225,
            cycle list={pattern=horizontal lines, , pattern=crosshatch},
        ]
        \addplot table[x=interval,y=Baseline,pattern=north east lines]{\thruvhundred};
        \addplot table[x=interval,y=Checkpoint]{\thruvhundred};
        \addplot table[x=interval,y=Tempo]{\thruvhundred};
        \node (1.03x) at (axis cs:128,210) {\scriptsize \mybox{1.03$\times$}};
        \node (1.04x) at (axis cs:512,70) {\scriptsize \mybox{1.04$\times$}};
        \legend{}
    \end{axis}
\end{tikzpicture}
    \newsubcap{V100\hspace*{65pt}} \label{fig:M2}  
\end{subfigure}
\caption{Throughput experiments at the maximum batch size annotated
with the speedup \emph{over the best baseline}.}
\label{fig:thru_results}
\end{figure}

\paragraph*{Impact on Loss and Accuracy}

\begin{figure}[!htbp]
\begin{subfigure}{0.495\columnwidth}
    \centering
    \includegraphics[width=\columnwidth]{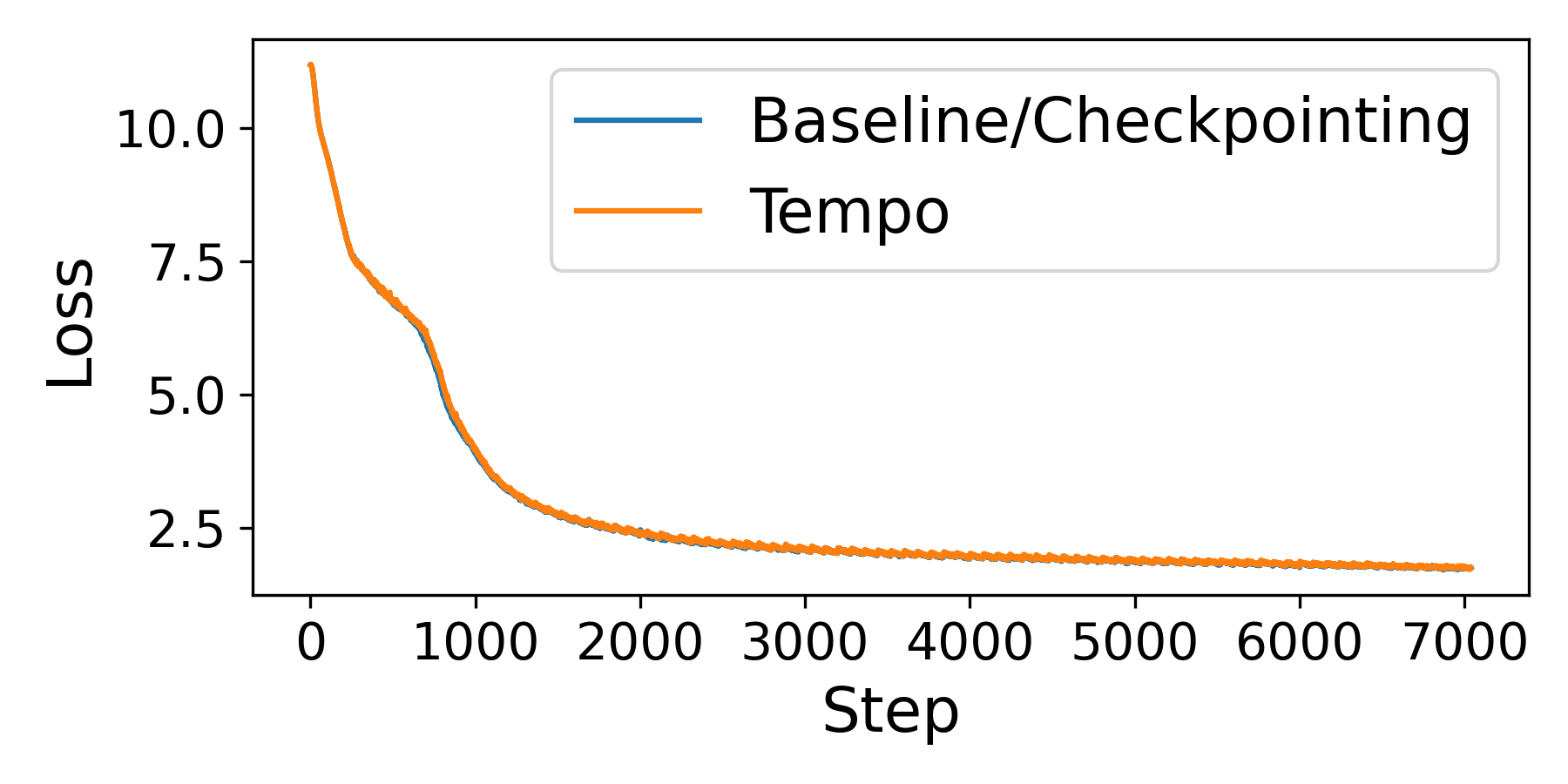}
    \newsubcap{Phase 1 BERT$_{BASE}$ pre-training curve on the English Wikipedia dataset \cite{wikipedia}.}
    \label{fig:loss_time}
\end{subfigure}
\begin{subfigure}{0.495\columnwidth}
    \centering
    \includegraphics[width=\columnwidth]{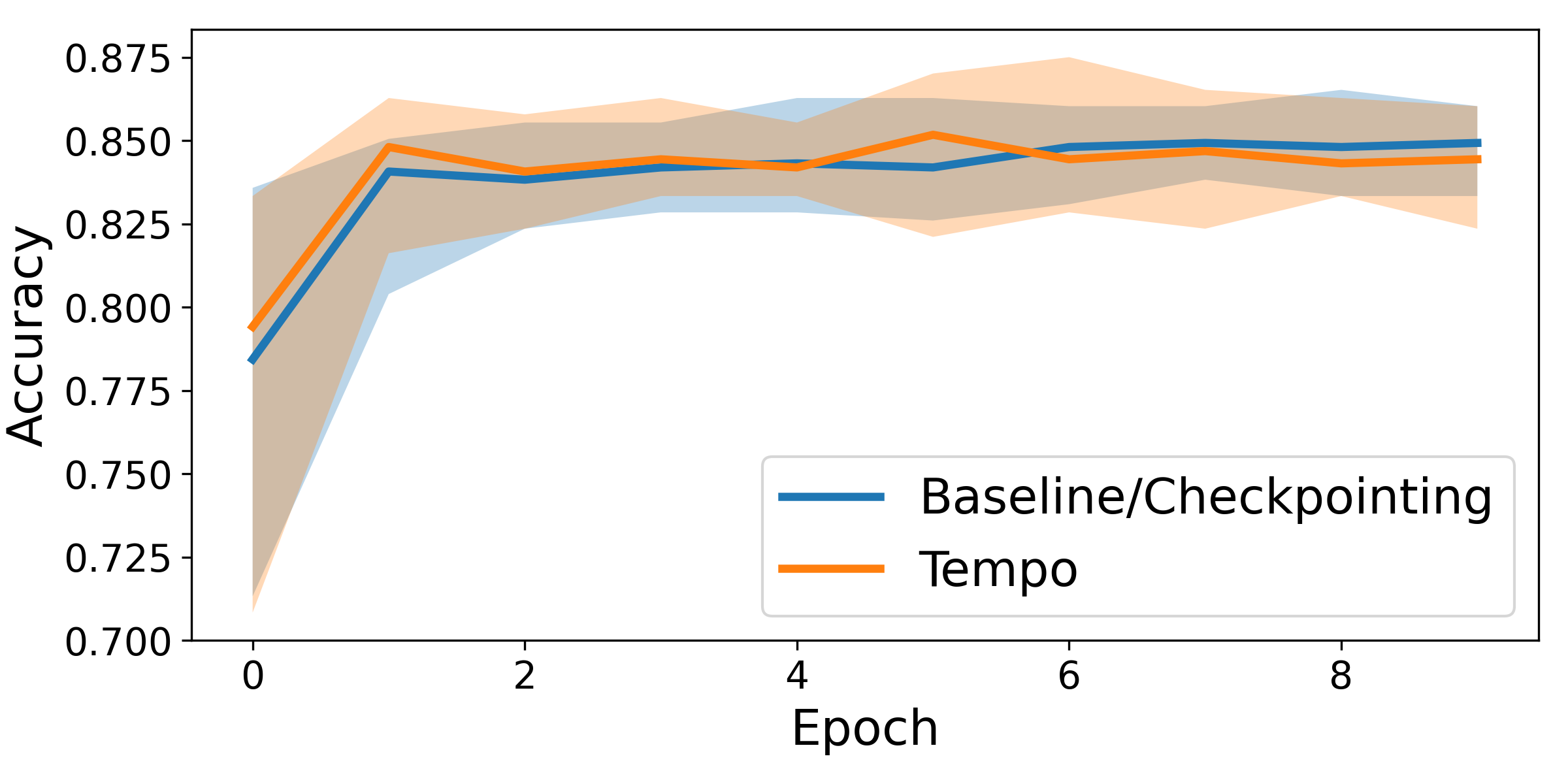}
    \newsubcap{Accuracy of BERT$_{BASE}$ fine-tuning \cite{devlin2018} on the MRPC \cite{mrpc} task. We run 10 trials of 10 epochs. The solid line represents the median accuracy of these trials, and the maximum and minimum along the training process by the transparent curves' boundaries.}
    \label{fig:accuracy_iteration}
\end{subfigure}
\end{figure}

We pre-train BERT$_{BASE}$ to ensure that our model's loss curve is not affected by approximate optimizations (e.g., In-Place GELU).
Figure~\ref{fig:loss_time} shows the loss curve of phase 1 of BERT$_{BASE}$ pre-training \cite{devlin2018}. We observe almost complete overlap in the loss curves with no more than a \emph{0.5\%} difference between \emph{Tempo} and the baseline at the endpoint. We conclude that within that margin of error our method is satisfactory.

For the fine-tuning accuracy, we use the pre-trained Huggingface \cite{huggingface} implementation. 
Figure~\ref{fig:accuracy_iteration} shows the results of BERT$_{LARGE}$ fine-tuning \cite{devlin2018} on the MRPC \cite{mrpc} task. The figure shows a consistent overlap between the maximum and minimum accuracy of \emph{Tempo} and the baseline, so we can conclude that \emph{Tempo} has little impact on the accuracy of the trained model.

\subsection{Ablation Studies}

\paragraph*{Ablation Study With Respect to Larger Model Parameters on Modern Hardware Platforms}

\begin{figure}[!htbp]  
\centering 
  \begin{subfigure}[b]{0.245\columnwidth}
    \begin{tikzpicture}
    \begin{axis}[
            ybar,
            xlabel=,
            bar width=.1cm,
            width=1.3\columnwidth,
            height=1.75\textwidth,
            symbolic x coords={128, 512},
            enlarge x limits=0.75,
            xtick=data,
            ymin=0,ymax=1.5,
            ylabel style={align=center}, ylabel={Normalized Throughput},
            cycle list={pattern=horizontal lines, , pattern=crosshatch},
        ]
        \addplot+[ybar] plot coordinates {(128, 1) (512, 1)};
        \addplot+[ybar] plot coordinates {(128, 1.02) (512, 1.09)};
        \addplot+[ybar] plot coordinates {(128, 1.09) (512, 1.16)};
        \node (1.07x) at (axis cs:128,1.35) {\tiny \mybox{1.07$\times$}};
        \node (1.07x) at (axis cs:512,1.35) {\tiny \mybox{1.07$\times$}};
        \legend{}
    \end{axis}
\end{tikzpicture}
   \end{subfigure}
\hspace*{11pt}
\begin{subfigure}[b]{0.245\columnwidth}
    \begin{tikzpicture}
    \begin{axis}[
            ybar,
            xlabel=,
            bar width=.1cm,
            width=1.3\columnwidth,
            height=1.75\textwidth,
            symbolic x coords={128, 512},
            enlarge x limits=0.75,
            xtick=data,
            yticklabels={,,},
            ymin=0,ymax=1.5,
            cycle list={pattern=horizontal lines, , pattern=crosshatch},
        ]
        \addplot+[ybar] plot coordinates {(128, 1) (512, 1)};
        \addplot+[ybar] plot coordinates {(128, 0.87) (512, 0.92)};
        \addplot+[ybar] plot coordinates {(128, 1.04) (512, 1.11)};
        \node (1.04x) at (axis cs:128,1.35) {\tiny \mybox{1.04$\times$}};
        \node (1.11x) at (axis cs:512,1.35) {\tiny \mybox{1.11$\times$}};
        \legend{};
    \end{axis}
\end{tikzpicture}
\end{subfigure}
\hspace*{-18pt}
\begin{subfigure}[b]{0.245\columnwidth}
    \begin{tikzpicture}
    \begin{axis}[
            ybar,
            xlabel=,
            bar width=.1cm,
            width=1.3\columnwidth,
            height=1.75\textwidth,
            symbolic x coords={128, 512},
            enlarge x limits=0.75,
            xtick=data,
            yticklabels={,,},
            ymin=0,ymax=1.5,
            cycle list={pattern=horizontal lines, , pattern=crosshatch},
        ]
        \addplot+[ybar] plot coordinates {(128, 1) (512, 1)};
        \addplot+[ybar] plot coordinates {(128, 1.15) (512, 1.35)};
        \addplot+[ybar] plot coordinates {(128, 1.17) (512, 1.32)};
        \node (1.02x) at (axis cs:128,1.35) {\tiny \mybox{1.02$\times$}};
        \node (0.98x) at (axis cs:512,1.45) {\tiny \mybox{0.98$\times$}};
        \legend{};
    \end{axis}
\end{tikzpicture}
\end{subfigure}
\hspace*{-18pt}
\begin{subfigure}[b]{0.245\columnwidth}
    \begin{tikzpicture}
    \begin{axis}[
            ybar,
            xlabel=,
            bar width=.1cm,
            width=1.3\columnwidth,
            height=1.75\textwidth,
            legend style={at={(1.2,0.85)},
                anchor=north,legend columns=3},
            symbolic x coords={128, 512},
            enlarge x limits=0.75,
            xtick=data,
            yticklabels={,,},
            ymin=0,ymax=1.5,
            cycle list={pattern=horizontal lines, , pattern=crosshatch},
        ]
        \addplot+[ybar] plot coordinates {(128, 1) (512, 1)};
        \addplot+[ybar] plot coordinates {(128, 1.01) (512, 1.20)};
        \addplot+[ybar] plot coordinates {(128, 1.16) (512, 1.39)};
        \node (1.15x) at (axis cs:128,1.35) {\tiny \mybox{1.15$\times$}};
        \node (1.16x) at (axis cs:512,1.45) {\tiny \mybox{1.16$\times$}};
    \end{axis}
\end{tikzpicture}
\end{subfigure}
\hspace*{139pt}
\begin{subfigure}[b]{\columnwidth}
    \begin{tikzpicture} 
    \begin{axis}[%
    hide axis,
    xmin=50,
    xmax=100,
    ymin=0,
    ymax=0.4,
    legend style={draw=white!15!black,legend cell align=left, legend columns=3, label=above : {Sequence Length}, at={(3, 3)}}
    ]
    \addlegendimage{pattern=horizontal lines,ybar,ybar legend}
    \addlegendentry{Baseline};
    \addlegendimage{pattern=none,ybar,ybar legend}
    \addlegendentry{Checkpoint};
    \addlegendimage{pattern=crosshatch,ybar,ybar legend}
    \addlegendentry{Tempo};
    \end{axis}
\end{tikzpicture}
\end{subfigure}
\caption{Normalized throughput at the maximum batch size, with annotated speedup \emph{over the best baseline}. From left to right the configurations are (a) BERT$_{LARGE}$ ($H = 1024$), (b) BERT$_{BASE}$ $H = 2048$, (c) BERT$_{LARGE}$ $H = 2048$, (d) BERT$_{BASE}$ $H = 3072$.}
\label{fig:thru_hidden_size}
\end{figure}
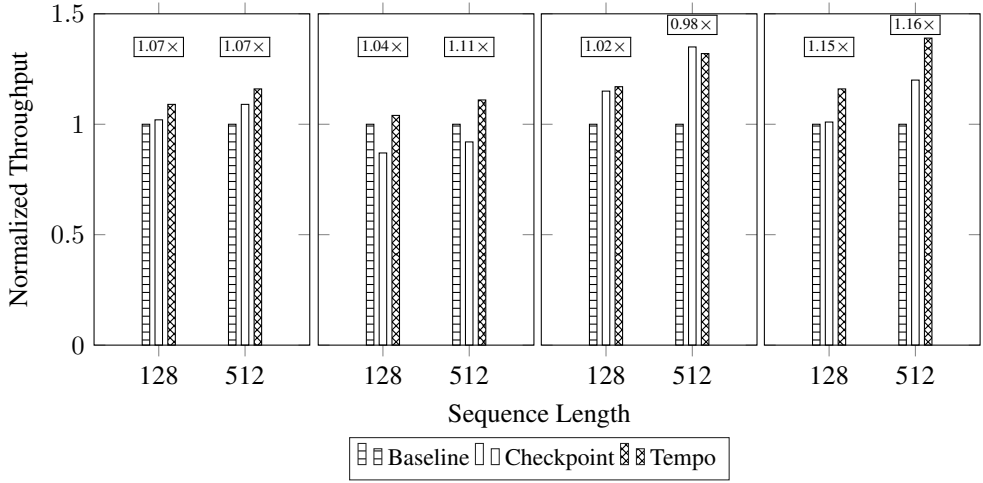

We also evaluate on other hardware platforms as well as model parameters. First, we use an increased hidden layer size for various configurations. These experiments are conducted on a platform with an NVIDIA A100 GPU \cite{a100} across sequence lengths of 128 and 512. We maintain the hidden layer size \(H\) to the number of attention heads \(A\) ratio of 64 which is in line with prior works \cite{vaswani2017, devlin2018}. The results are shown in Figure \ref{fig:thru_hidden_size}. The figure demonstrates two important generalizations of \emph{Tempo}. First, note that even on newer and more advanced GPUs, \emph{Tempo} continues to provide a tangible benefit. Second, across larger hidden layer sizes \emph{Tempo} consistently demonstrates a clear improvement over the baseline (as shown in the figure, this can be as high as a \textbf{39\% speedup} over \emph{Baseline} which corresponds to a 16\% speedup over \emph{Checkpoint}). The speedup over \emph{Checkpoint} is as high as \textbf{20\%} . We conclude that \emph{Tempo} will continue to be applicable to new hardware and larger models. 

\begin{figure}[!htbp]  
\centering 
    \begin{tikzpicture}
    \begin{axis}[
            ybar,
            bar width=.15cm,
            width=0.7\columnwidth,
            height=0.28\textwidth,
            legend style={at={(1.2,0.85)},
                anchor=north,legend columns=1},
            symbolic x coords={128, 512, 1024, 2048, 3072},
            enlarge x limits=0.2,
            xtick=data,
            ymin=0,ymax=1.5,
            ylabel style={align=center}, ylabel={Normalized\\Throughput},
            cycle list={pattern=horizontal lines, , pattern=crosshatch},
        ]
        \addplot+[ybar] plot coordinates {(128, 1) (512, 1) (1024, 1) (2048, 1) (3072, 0)};
        \addplot+[ybar] plot coordinates {(128, 0.90) (512, 0.91) (1024, 1.09) (2048, 1.09) (3072, 1)};
        \addplot+[ybar] plot coordinates {(128, 1.06) (512, 1.09) (1024, 1.27) (2048, 1.24) (3072, 1.10)};
        \node (1.06x) at (axis cs:128,1.35) {\scriptsize \mybox{1.06$\times$}};
        \node (1.10x) at (axis cs:512,1.35) {\scriptsize \mybox{1.10$\times$}};
        \node (1.16x) at (axis cs:1024,1.4) {\scriptsize \mybox{1.16$\times$}};
        \node (1.16x) at (axis cs:2048,1.4) {\scriptsize \mybox{1.16$\times$}};
        \node (1.10x) at (axis cs:3072,1.35) {\scriptsize \mybox{1.10$\times$}};
        \legend{Base., Chk., Tempo}
    \end{axis}
\end{tikzpicture}
\caption{Normalized throughput relative to the Baseline across different sequence lengths on the NVIDIA A100 GPU for BERT$_{LARGE}$ modified to use 12 layers. We annotate each bar group with \emph{Tempo's} speedup \emph{over the best baseline}.}
\label{fig:sequence_length}
\end{figure}
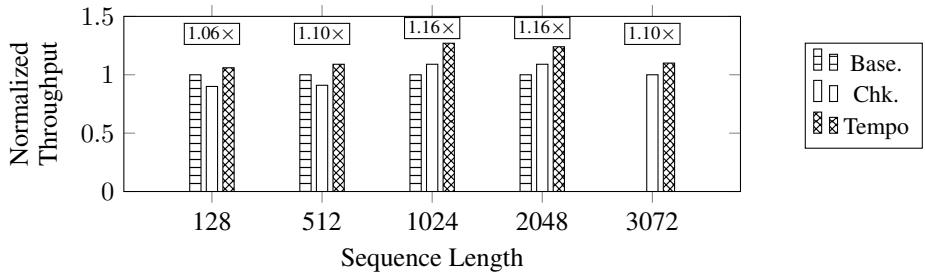

We also conduct experiments on BERT$_{LARGE}$ (modified to use 12 Layers instead of 24 for more data points) for sequence lengths larger than 512. Figure~\ref{fig:sequence_length} shows the results for this experiment, where we demonstrate that \emph{Tempo} outperforms \emph{Baseline} on longer sequence lengths as well which can be as high as a \textbf{27\% speedup} over \emph{Baseline}. At the same settings, we observe 16\% speedup over \emph{Checkpoint}. Tempo also outperforms \emph{Checkpoint} by as much as \textbf{20\%}. We conclude that yet again \emph{Tempo} will be able to take advantage of modern hardware, as well as remain advantageous as sequence lengths increase. Note that the largest sequence length of 3072 on \emph{Baseline} does not have enough memory to run.

\paragraph*{Results on Other Models}

We conduct experiments on other Transformer-based models as well: RoBERTa \cite{liu2019roberta} and GPT2 \cite{gpt2}. For the evaluation of RoBERTa, we use the Fairseq library \cite{ott2019fairseq} as well as a sequence length of 512, while GPT2 uses the Huggingface GPT2 model \cite{huggingface}. These experiments are conducted on both 2080 Ti and V100 setups. We note that the improvement over the baseline is substantial (up to \textbf{19\% and 26\%} for GPT2 and RoBERTa respectively on the 2080 Ti setup. This improvement corresponds to a increase in batch size of \textbf{3$\times$} and \textbf{2$\times$}. Furthermore, we also see speedups of 5\% and 4\% on the V100 setup as well. From these results, we conclude that \emph{Tempo} generalizes well to other Transformer-based models besides BERT.

\goodbreak\noindent

\section{Extensions}
\label{sec:Extensions}

\subsection{Extending In-place GELU}

The ideas used in section \ref{sec:Key_Ideas} for In-place GELU can be extended to general elementwise layers. The generic steps required for this are listed below, from the the high-level mathematics to the low-level kernel based accelerator implementation. To the best of our knowledge, ours is the first work that exposes this potential optimization. This is a generic strategy to reduce memory footprint in a multi-dimensional space.

Consider an elementwise layer with $n$ inputs that applies a function $f$ inputs such that $y = f(x_1, x_2,...,x_n)$ to each corresponding element of the input tensors and where the output is retained for the backward pass of the subsequent layer.

    \textbullet \hspace{1em} Discard activation $x_1$ without loss of generality. Determine a function $g$ such that $x_1 = g(y, x_2, ..., x_n)$. For bijective functions of one variable this is simply the inverse. \newline
    \textbullet \hspace{1em} If such a function does not exist without ambiguity, construct functions $g_1 ,..., g_j$ that can recover $x_1$ on an interval. Construct a function $g_*$ such that $x_1 = g(m, y, x_2, ... x_n)$ where $m$ is an indicator that denotes the interval from which $x_1$ from and thus the piecewise selection of one of $g_1, ..., g_j$.  Polynomially approximate each of $g_1, ..., g_j$ to construct a new piecewise function $g_{*p}$ in the case that they cannot be expressed analytically. \newline
    \textbullet \hspace{1em} For the implementation of the forward pass, fold the computation of $m$ into the computation of $f$. In essence, construct a new function $f_*$ such that $(y, m) = f_*(x_1, ..., x_n)$. This can be done in a single kernel call. \newline
    \textbullet \hspace{1em} For the backward pass, fold the calculation of $x_1 = g_{*p}(m, y, x_2, ..., x_n)$ into the computations of $\frac{\partial f}{\partial x_2}, ..., \frac{\partial f}{\partial x_n}$ if the computation of these values requires $x_1$ by composing these functions. In essence, fusing the kernels for the inverse and gradient operator. Then, we require $n$ kernel calls to calculate the gradient with respect to the loss as before.

We illustrate this strategy in appendix \ref{appdix:FirstImplDetails} in more detail. The crux of the idea is that $m$, if needed at all, can be stored with less memory than $x_1$, while keeping the number of kernel calls to a minimum.

\subsection{Auto-Tempo}

As part of exploratory future work, we consider the application of \textit{Tempo} as an automatic compiler pass. We propose and prototype two different methods of automatically applying \textit{Tempo} to transformers which are available at the link in section \ref{sec:Conclusions}. The first method is a fast method of profiling beforehand to determine whether memory footprint reduction would help, then applying \textit{Tempo} to all applicable layers. The second method is a fine-grained method applies \textit{Tempo} to a subset of the applicable layers where the subset is determined through automatic profile and search, analogous to binary search.
\vspace{-0.2cm}
\section{Conclusion}
\label{sec:Conclusions}
We propose \textit{Tempo}, a new mechanism that reduces the memory footprint of Transformer-based models at low cost. It shows an improvement in throughput of up to \maxthru over the state-of-the-art baseline for BERT$_{LARGE}$ pre-training  task and also shows an improvement in maximum batch size of up to \maxbatchsize on both V100 and 2080 Ti GPUs. Our technique also generalizes well to new models, more modern hardware, as well as diverse model parameters in terms of memory footprint and throughput, demonstrating the robustness of our technique. Our hope is that \textit{Tempo} will be used with other footprint reduction methods to improve training efficiency of Transformer-based models. We open-source \emph{Tempo} for an immediate positive impact on both machine learning researchers and practitioners here: \url{https://github.com/UofT-EcoSystem/Tempo}.
\clearpage
\bibliographystyle{plain}
\bibliography{biblio.bib}

\clearpage

\appendix

\section*{Summary of Appendices}

These appendices contain additional details to cover different aspects of our work: motivation, related work, implementation, methodology, and extra ablation studies.

Appendix \ref{sec:memory} contains a more detailed breakdown of the memory usage of BERT$_{BASE}$.

Appendix \ref{sec:related} goes over relevant related works and the difference between them and our work.

Appendix \ref{sec:methods} goes more in depth into the major memory reduction footprint techniques, and compare them to our method.

Appendix \ref{appdix:Layernorm} covers details of the LayerNorm backward pass derivation.

Appendix \ref{appdix:FirstImplDetails} includes implementation details of our optimizations.

Appendix \ref{appdix:ImplDetails} includes additional information on the implementation of our optimizations.

Appendix \ref{appdix:Setup} includes a more detailed description of our experimental setups.

Appendix \ref{appdix:ablation} includes a memory footprint ablation study with respect to Tempo optimizations.

Appendix \ref{appdix:checklist} is the NeurIPS paper checklist.

\section{A Closer Look at the Memory Breakdown of \texorpdfstring{BERT$_{BASE}$}{BERT Base}}
\label{sec:memory}

\begin{figure}[!htbp]
    \centering
    \includegraphics[width=\columnwidth]{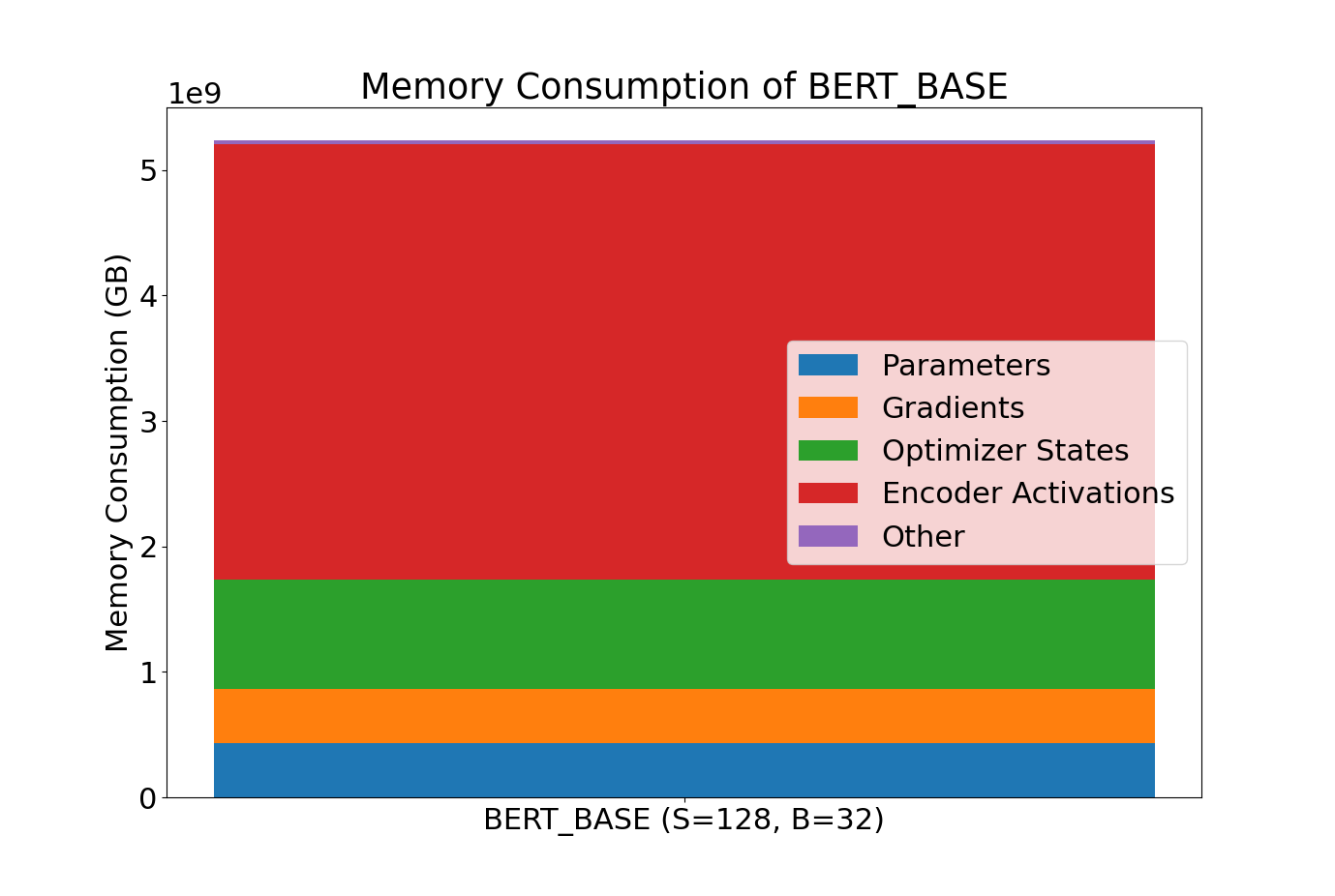}    
    \caption{GPU memory breakdown for BERT$_{BASE}$ \cite{devlin2018} fine-tuning on the MRPC \cite{mrpc} task using the Huggingface \cite{huggingface} at a sequence length of 128 and batch size of 32.}
    \label{fig:memory_breakdown}
\end{figure}

Figure \ref{fig:memory_breakdown} shows a detailed memory breakdown of the Huggingface BERT$_{BASE}$ implementation \cite{huggingface} on the MRPC \cite{mrpc} fine-tuning task at a batch size of 32, profiled using the skyline tool \cite{skyline}. From the figure, encoder layer activations are clearly the major contributor to the memory footprint compared to parameter weights, gradients, and optimizer states.

\section{Related Works}
\label{sec:related}

There has been a number of previous works~\cite{Sanh2019DistilBERTAD, jiao-etal-2020-tinybert, longformer, performer, big-bird} which focus on developing models that remain competitive with BERT and other Transformer-based models while requiring less memory and compute. One approach that has been taken is to tackle the $\mathcal{O}(S^2)$ nature of the attention mechanism \cite{survey}. Out of these, a few interesting and relevant ideas in terms of memory footprint reduction involve sparsifying the attention mechanism \cite{longformer, big-bird}, or using a decomposable softmax layer in order to avoid doing the $\mathcal{O}(S^2)$ computation in the first place \cite{performer}. In contrast, our technique targets a broad spectrum of sequence lengths. Other models such as TinyBERT \cite{jiao-etal-2020-tinybert} and DistilBERT \cite{Sanh2019DistilBERTAD} take a different approach in reducing the total model size. In these works, the technique of \textit{Distillation} is used to train a smaller student network from a larger teacher network \cite{HinVin15Distilling}. These approaches are model level algorithm changes, and therefore orthogonal to our work.

Two other important techniques are mixed precision training~\cite{mixed} and in-place activated BatchNorm~\cite{rotabulo2017}. Mixed precision training involves training using both 32-bit and 16-bit IEEE floating point numbers depending on the numerical sensitivity of different layers \cite{mixed}. This both decreases the training time and the memory pressure of the model dependent on the hardware \cite{mixed}. Reversible BatchNorm uses techniques designed to optimize the memory footprint of CNN models which make extensive use of RELU and BatchNorm operators \cite{rotabulo2017}. They do this by deriving in-place expressions for these operators (the input feature map for these operators no longer needs to be stored). Although this work shows better performance than checkpointing, it is also specific to CNN models, so is not directly comparable to our work. Other recent works such as Substation~\cite{datamovement} aims to improve training performance by reducing the total movement of data with operator fusion; this is also orthogonal to our memory footprint based approach.

Other techniques are more focused on inference and model weights, both of which are not applicable in this context of Transformer-based model \emph{training}, where the \emph{activations} are the memory bottleneck \cite{deepcompression, eie, eyeriss, conciseload, cnvlutin, diannao, stripes, scnn, scaledeep}.

\section{General Memory Footprint Reduction Techniques}
\label{sec:methods}

We expand on the techniques we referred to in section \ref{sec:Background_and_Motivation} of checkpointing, offloading, and compression in this section. As previously iterated in section \ref{sec:Background_and_Motivation}, (1) these techniques do not look closely at the specific structure of the BERT model, (2) At a per layer level, our techniques can provide better performance at a cheaper computational cost, and (3) these techniques are orthogonal to our work.

\subsection{Checkpointing}

Early work in this direction was able to reduce the memory cost of linear models to $\mathcal{O}(\sqrt{n})$ for general linear models in the number of layers. Further work  has expanded on this. Echo has innovative optimizations for specific models \cite{zheng2020}, Checkmate computes an optimal checkpointing schedule minimizing recomputation time with respect to a specific memory budget \cite{checkmate}, and Dynamic Tensor Rematerialization uses online heuristics to minimize recomputation time while training \cite{dtr}. Furthermore, there are works in this area that combine checkpointing with other techniques such as offloading detailed below.

\subsection{Offload}

Initial work in this direction such as vDNN focused on specific schemes to offload layer outputs depending on their computation cost \cite{vdnn}. There was also some consideration into pre-fetching feature maps in anticipation of their use in the backward pass. More work in that direction focuses on several different directions. For one, Superneurons \cite{superneurons} combines both checkpointing and offloading. In this work, they checkpoint computationally cheap layers, avoiding transfer overhead, while simultaneously offloading computationally expensive layers, avoiding recomputation overhead for those cases. Capuchin takes this further in considering tensor level accesses, as well as considering the runtime fetching and computation time in making decisions on which strategy to apply \cite{capuchin}. Additionally, ZeRO-Infinity combines offloading of feature maps with offloading of model states and other optimizations \cite{zeroinf}.

\subsection{Compression}

Works like Gist \cite{jain2018} and ActNN \cite{actnn} both include forms of lossy compression. Gist specifically targets CNN based models with a variety of different lossy and lossless optimizations, taking an approach similar to our own in examining the model structure closely \cite{jain2018}. However, as this is a work focused on CNNs, the techniques described do not directly apply to the BERT model we are aiming to optimize for. ActNN on the other hand is a more general technique, using a quantization strategy designed to reduce the number of bits needed for feature map storage, while preserving certain theoretical guarantees regarding model conversion \cite{actnn}. It additionally shows good performance relative to techniques such as Dynamic Tensor Rematerialization \cite{dtr} and Capuchin \cite{capuchin}. This work does not consider the BERT model however \cite{actnn}.

\section{Backward pass of In-place LayerNorm}
\label{appdix:Layernorm}

In the following, we show how to compute the gradients of LayerNorm layer using the output. 

\subsection{Notations}

We use $x$, $\hat{x}$, $y$, $\mu$ and $\sigma^2$ to represent the input, intermediate normalized input, output, and mean and variance of the input, respectively. The parameters of LayerNorm function, scaling factor and bias, are denoted by $\gamma$ and $\beta$, respectively. $L$ represents the loss.
For simplicity but without the loss of generality, we assume the size of input is $(N, M)$, where the second dimension represents all the dimensions that are needed to be normalized. The meanings, definitions and sizes of variables are listed in Table~\ref{tab:layernorm_notation}.
~\\

\begin{table}[!htbp]
    \centering
    \small
    \begin{tabular}{rcc}
        \toprule
        Meaning & Definition & Size  \\
        \midrule
        input & $x = \{ x_{ij}, i = 1,...,N, \ j=1,...,M\}$ & $(N, M)$ \\
        norm-input & $\hat{x} = \{ \hat{x}_{ij}, i = 1,...,N, \ j=1,...,M\}$ & $(N, M)$\\
        output & $y = \{ y_{ij}, i = 1,...,N, \ j=1,...,M\}$ & $(N, M)$ \\
        mean & $\mu = \{\mu_i, i = 1,...,N\}$ & $(N)$ \\
        variance & $\sigma^2 = \{\sigma_i^2, i = 1, ..., N\}$ & $(N)$  \\
        weight & $\gamma = \{\gamma_j, j = 1, ..., M\}$ & $(M)$   \\
        bias & $\beta = \{\beta_j, j = 1, ..., M\}$ & $(M)$    \\
        \bottomrule
    \end{tabular}
    \caption{The meanings, definitions and sizes of variables used in LayerNorm layer.}
    \label{tab:layernorm_notation}
\end{table}

\subsection{Forward Pass}

In the forward pass, input is firstly normalized along the second dimension, and then scaled and shifted accordingly.
\begin{align*}
    \hat{x}_{ij} &= \frac{x_{ij}-\mu_{i}}{\sqrt{\sigma_{i}^{2}+\epsilon}} \\
    y_{ij} &= \gamma_j \hat{x}_{ij}  + \beta_j
\end{align*}
where $\mu_i = \frac{1}{M} \sum_{j=1}^M x_{ij}$ and $\sigma_i^2 = \frac{1}{M} \sum_{j=1}^M (x_{ij} - \mu_{i})^2$. $\epsilon$ is added for numerical stability.

\subsection{Backward Pass}

Our goal is to use output to compute the gradients with minimum overhead. Intuitively however, the input is needed to compute the gradients of LayerNorm, which means we need to compute backwards to get input. We find that we can use the intermediate normalized input to get what we want. The gradient derivations are listed as follows, along the lines of the BatchNorm derivation in \cite{rotabulo2017}.

\begin{equation*}
    \frac{\partial y_{ij}}{\partial \gamma_j}=\hat{x}_{ij}, \quad \frac{\partial y_{ij}}{\partial \beta_j}=1, \quad \frac{\partial y_{ij}}{\partial \hat{x}_{ij}}=\gamma_j,
\end{equation*}
\begin{align*}
    \frac{\partial L}{\partial \gamma_j} &= \sum_{i=1}^{N} \frac{\partial L}{\partial y_{ij}} \frac{\partial y_{ij}}{\partial \gamma_{j}} = \sum_{i=1}^{N} \frac{\partial L}{\partial y_{ij}} \hat{x}_{ij},   \\
    \frac{\partial L}{\partial \beta_j} &= \sum_{i=1}^{N} \frac{\partial L}{\partial y_{ij}} \frac{\partial y_{ij}}{\partial \beta_{j}} = \sum_{i=1}^{N} \frac{\partial L}{\partial y_{ij}}, \\
    \frac{\partial L}{\partial \hat{x}_{ij}} &=\frac{\partial L}{\partial y_{ij}} \frac{\partial y_{ij}}{\partial \hat{x}_{ij}} =\frac{\partial L}{\partial y_{ij}} \gamma_j,
\end{align*}
Here we can the gradients with regard to $\gamma$, $\beta$ and $\hat{x}$. We still need to derive the gradient to input further.
\begin{equation*}
    \frac{\partial \hat{x}_{ij}}{\partial \sigma_{i}^2} = - \frac{\hat{x}_{ij}}{2 (\sigma_i^2 + \epsilon)}, \quad \frac{\partial \hat{x}_{ij}}{\partial \mu_{i}^2} =  - \frac{1}{\sqrt{\sigma_i^2 + \epsilon}},
\end{equation*}
~
\begin{equation*}
    \begin{aligned}
        \frac{\partial L}{\partial \sigma_i^2} &= \sum_{p=1}^{N} \sum_{q=1}^M \frac{\partial L}{\partial \hat{x}_{pq}} \frac{\partial \hat{x}_{pq}}{\partial \sigma_i^2} \\
        &= \sum_{q=1}^M \frac{\partial L}{\partial \hat{x}_{iq}} \frac{\partial \hat{x}_{iq}}{\partial \sigma_i^2}	\quad (p = i)	\\
        &= \sum_{j=1}^M \frac{\partial L}{\partial \hat{x}_{ij}} \frac{\partial \hat{x}_{ij}}{\partial \sigma_i^2}	\quad (let\  q = j)	\\
        &= \sum_{j=1}^M \frac{\partial L}{\partial y_{ij}} \gamma_j \cdot \left( - \frac{\hat{x}_{ij}}{2 (\sigma_i^2 + \epsilon)} \right),
    \end{aligned}
\end{equation*}
~
\begin{equation*}
    \begin{aligned}
        \frac{\partial L}{\partial \mu_i} &= \sum_{p=1}^{N} \sum_{q=1}^M \frac{\partial L}{\partial \hat{x}_{pq}} \frac{\partial \hat{x}_{pq}}{\partial \mu_i} \\
        &= \sum_{q=1}^M \frac{\partial L}{\partial \hat{x}_{iq}} \frac{\partial \hat{x}_{iq}}{\partial \mu_i}	\quad (p = i)	\\
        &= \sum_{j=1}^M \frac{\partial L}{\partial \hat{x}_{ij}} \frac{\partial \hat{x}_{ij}}{\partial \mu_i}	\quad (let\  q = j)	\\
        &= \sum_{j=1}^M \frac{\partial L}{\partial y_{ij}} \gamma_j \cdot \left( - \frac{1}{\sqrt{\sigma_i^2 + \epsilon}} \right),
    \end{aligned}
\end{equation*}
~
\begin{equation*}
    \frac{\partial \sigma_i^2}{\partial x_{ij}} = \frac{2 (x_{ij} - \mu_i)}{M}, \quad \frac{\partial \mu_{i}}{\partial x_{ij}}=\frac{1}{M}, \quad \frac{\partial \hat{x}_{ij}}{\partial x_{ij}}=\frac{1}{\sqrt{\sigma_{i}^{2}+\epsilon}},
\end{equation*}
Combining all the intermediate results above, we have
\begin{equation*}
    \begin{aligned}
        \frac{\partial L}{\partial x_{ij}} &= \sum_{p=1}^{N} \sum_{q=1}^M \left( \frac{\partial L}{\partial \hat{x}_{pq}} \frac{\partial \hat{x}_{pq}}{\partial x_{ij}}\right) + \sum_{p=1}^N \left( \frac{\partial L}{\partial \sigma_p^2} \frac{\partial \sigma_p^2}{\partial x_{ij}} + \frac{\partial L}{\partial \mu_p} \frac{\partial \mu_p}{\partial x_{ij}}\right)\\
        &= \frac{\partial L}{\partial \hat{x}_{ij}} \frac{\partial \hat{x}_{ij}}{\partial x_{ij}} + \frac{\partial L}{\partial \sigma_i^2} \frac{\partial \sigma_i^2}{\partial x_{ij}} + \frac{\partial L}{\partial \mu_i} \frac{\partial \mu_i}{\partial x_{ij}}	\\
        &=\Bigg[ \frac{\partial L}{\partial y_{ij}} \gamma_j - \bigg( \sum_{j=1}^m \frac{\partial L}{\partial y_{ij}} \gamma_j \cdot \hat{x}_{ij} \bigg) \cdot \frac{\hat{x}_{ij}}{m} \\
         & \qquad - \bigg( \sum_{j=1}^m \frac{\partial L}{\partial y_{ij}} \gamma_j \bigg) \cdot \frac{1}{m}\Bigg] \cdot \frac{1}{\sqrt{\sigma_i^2 + \epsilon}}
    \end{aligned}
\end{equation*}
where intermediate normalized input $\hat{x}$ can be computed as $    \hat{x}_{ij} = (y_{ij} - \beta_j) / \gamma_j$. Therefore, by extra stashing weight (scaling factor) $\gamma$, bias $\beta$ and variance of the input $\gamma^2$, we can get the gradients using output without recovering input.

\section{Implementation Details}

\label{appdix:FirstImplDetails}

\subsection{In-Place GELU}
    As we show in Section~\ref{sec:Key_Ideas}, it is possible to compute the inverse of the GELU function by knowing what side of the minimum the input originated from. This is shown in Equation \ref{eq:gelu_inverse} where $\operatorname{GELU*}$ is a function that returns both the $\operatorname{GELU}$ output and $m$ is the mask bit that denotes which side of the minimum the input originates from.
    \begin{equation} \label{eq:gelu_inverse}
        \operatorname{GELU*}^{-1} (\operatorname{GELU}(x), m) = x
    \end{equation}
    
\begin{figure}[!htbp]
    \centering
    \begin{subfigure}[b]{0.495\columnwidth}
        \centering
        \includegraphics[width=\columnwidth]{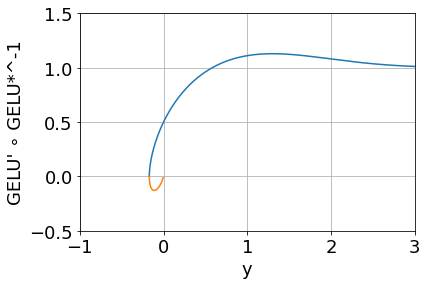}
        \caption{GELU derivative from \eqref{eq:gelu_compose}. The section corresponding to $x > -0.75179$ is in blue, $x \leq -0.75179$ -- in orange.}
        \label{fig:gelu_derivative}
    \end{subfigure}
    \begin{subfigure}[b]{0.495\columnwidth}
        \centering
         \includegraphics[width=\columnwidth]{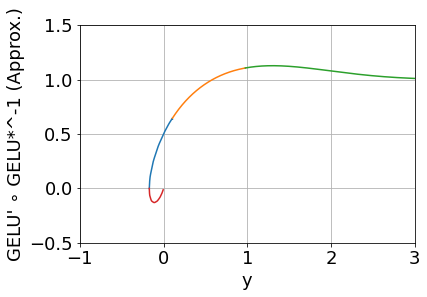}
        \caption{Our approximation of function \eqref{eq:gelu_compose} with a piece-wise polynomial approximation. Different sections of the approximation are shown in different colours.}
        \label{fig:ourgeluderiv}
    \end{subfigure}
    \end{figure}

   Moreover, we observe that we do not need to compute the inverse and then compute the derivative with respect to the input in a two step process as a na\"{i}ve approach would suggest. Instead, we can precompute this composition (see Equation \eqref{eq:gelu_compose}):
    \begin{equation} \label{eq:gelu_compose}
        \frac{d \operatorname{GELU}}{dx}(y) = \operatorname{GELU}' \circ \operatorname{GELU*}^{-1} (y, m)
    \end{equation}
    in order to \textbf{compute the derivative with respect to the input} directly using the output value. A plot of this relation is shown in Figure~\ref{fig:gelu_derivative}.

    A key observation about the GELU function is that it is transcendental, and hence there is no simple solution for the inverse of the GELU function in terms of elementary functions~\cite{transcendental}. We therefore approximate sections of Equation~\eqref{eq:gelu_compose} with a piece-wise polynomial with a degree up to 13.\footnote[5]{Additional details of this implementation are covered in Appendix \ref{appdix:ImplDetails}.} A plot of this approximation is shown in Figure~\ref{fig:ourgeluderiv}. 
    
\subsection{In-Place LayerNorm}
    As stated in Section \ref{sec:Key_Ideas}, we aim to reuse the output of the LayerNorm \cite{layernorm}, which must be stored for the backpropagation of the successive fully-connected layers, while discarding the input. Although prior work has demonstrated the usability of In-Place Activated BatchNorm in the context of CNN networks \cite{rotabulo2017}, we note that this approach is not applicable in the Transformer case, which employs LayerNorm instead~\cite{vaswani2017}.
    
    By employing an alternative derivation for the gradient of LayerNorm which stashes alternative parameters, we can compute the gradients with \emph{negligible} performance overhead while achieving ideal memory footprint reduction for this operator. Following a similar approach as for In-Place GELU, we implement this operator as a Python PyTorch module, allowing it to be easily substituted in place of the existing LayerNorm layer in an implementation of the BERT model~\cite{deeplearningexamples, huggingface}. See \ref{appdix:ImplDetails} for additional implementation details and the full derivation.
    
\subsection{Sub-Layer Dropout Recomputation}

    Our basic implementation of Sub-Layer Dropout Recomputation follows the example of our other optimizations. Note that the way Dropout is implemented requires a randomly generated mask, where a portion of the inputs are set to zero according to a percentage $p$, which is also needed in the backward pass~\cite{pytorch, mxnet}. We simply stash this mask, and discard the output. Then, in the backward pass, we recompute the output as shown in Figure \ref{fig:dropout}. Storing a boolean mask of size $N$ will take $N \times 1$ bytes, boolean tensors in PyTorch use 1 byte per value, while 32-bit floating points will use $N \times 4$ bytes~\cite{pytorch}. Therefore, the total memory saved by discarding the output will be $4/5$ of the total pre-optimization dropout output memory cost.
    
    In contrast to prior works which modify the framework and may not expose this optimization \cite{capuchin, dtr_implementation, checkmate_implementation}, we develop a PyTorch module which can be added in to reduce the memory pressure of the critical attention section of the Transformer-based models~\cite{vaswani2017} with minimal overhead. Appendix \ref{appdix:ImplDetails} provides more in-depth comparison between prior work and our implementation.

\subsection{Other Engineering Optimizations}
We note that PyTorch uses a memory-inefficient implementation of the softmax function which retains both the input and output of the function for the backward pass \cite{pytorch}. Instead, only the output is necessary. This optimization has also previously been implemented as part of some models in the Huggingface library \cite{deberta_huggingface}. We use this optimization as well in our implementation of the attention mechanism.

\subsection{Elementwise Extension}

\begin{figure}[!htbp]
    \centering
    \includegraphics[width=0.75\columnwidth]{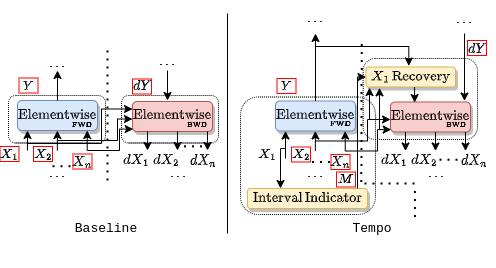}
    \caption{Comparison of our in-place general elementwise strategy with the baseline. Dotted border rectangles enclose functions that can be executed in a single kernel. Red bordered activations and gradients are needed for the computations of the elementwise and successive layer's backward pass.}
    \label{fig:elementwise}
\end{figure}

We illustrate the difference between our general elementwise strategy in Figure \ref{fig:elementwise}.

\section{Additional Implementation Details}
\label{appdix:ImplDetails}

This section contains additional implementation details of our technique.

\subsection{In-place GELU}

We implement this optimization in PyTorch. In this case, we write the forward pass to return the GELU of the input function, as well as a boolean mask indicating whether the input is greater than or equal to the point at which the minimum value occurs, $x \approx -0.75179$. The backward pass is slightly more complicated. We take as inputs to the backward pass the incoming gradient, the saved mask, as well as the saved output values. The corresponding approximating polynomial shown in Figure \ref{fig:ourgeluderiv} is determined and then computed. We use approximating polynomials of up to degree 13 in this case.

We implement the forward and backward pass using CUDA~\cite{cuda} which is wrapped in C++~\cite{C++}. Both of these are wrapped in a Python~\cite{python} PyTorch layer to be substituted easily for an existing implementation. 
While profiling our implementation we found that the memory latency of loading these inputs, as well as storing the output to be the bottleneck. We were thereby able to implement polynomials of degree 13, since the computation is hidden by the memory access latency, although better approximations may be possible.

\subsection{In-place LayerNorm}

 In-place Layernorm is implemented as a custom PyTorch module \cite{pytorch}. This is done in 3 stages as per PyTorch's custom module implementation. To do this, we write a custom CUDA \cite{cuda} implementation based on PyTorch's own implementation of the LayerNorm layer \cite{pytorch}. We write a forward and backward layer wrapper in C++ \cite{C++}, which is then again wrapped as a Python \cite{python} PyTorch module, allowing it to be easily substituted in place of the existing LayerNorm layer in an implementation of the BERT model such as the NVIDIA DeepLearningExamples BERT implementation or the Huggingface implementation \cite{deeplearningexamples, huggingface}.     
 
\subsection{Sub-Layer Dropout Recomputation}

We note that although the state-of-the-art recomputation/checkpointing papers~\cite{checkmate, dtr} consider such an abstract idea in theory, their practical implementations~\cite{checkmate_implementation, dtr_implementation} never treat the sub-layer granularity provided in the frameworks as the lowest granularity of those techniques' applicability. We concede that in the case of TensorFlow, the dropout layer has additional underlying granularity as a result of its implementation~\cite{tensorflow}, and hence this would not be applicable in that case -- and Checkmate \cite{checkmate} would consider this level of granularity. However, this is not the case in PyTorch's checkpointing implementation \cite{check_pytorch}. The tensor-level granularity of optimization is noted in Capuchin~\cite{capuchin} as well. However, Capuchin is a technique that requires runtime level profiling, and modifications to the framework. It may or may not offload, recompute, or otherwise store any part of the dropout layer specified.

Our method uses in-built PyTorch operators at a C++ level to rewrite the attention mechanism, which is again wrapped as a Python PyTorch module to be substituted.

\section{Experimental Setup}
\label{appdix:Setup}

Table \ref{tab:infra} shows a more detailed view of our experimental platform.

     \begin{table}[!htbp]
         \centering
         \scriptsize
         \begin{tabular}{cccccccc}
             \toprule
             \\[0.5in]
             \begin{rotate}{60} GPU \end{rotate} & 
             \begin{rotate}{60} \# GPUs \end{rotate} & 
             \begin{rotate}{60} GPU Mem. (GB) \end{rotate} &
             \begin{rotate}{60} CUDA Version \end{rotate} &
             \begin{rotate}{60} GPU Driver \end{rotate} &
             \begin{rotate}{60} Pytorch Version \end{rotate} &
             \begin{rotate}{60} \# (v)CPUs \end{rotate} &
             \begin{rotate}{60} Sys. Mem. (GiB) \end{rotate} \\
              \midrule
              2080 Ti & 4 & 11 & 11.2 & 460.27.04 & 1.9.0 & 64 & 126 \\
              V100 & 4 & 16 & 11.0 & 450.142.00 & 1.9.0 & 32 & 244 \\
              A100 & 1 & 40 & 11.2 & 460.32.03 & 1.9.0 & 64 & 250 \\
              \bottomrule
         \end{tabular}
         \caption{A short summary of our test setups, including CUDA \cite{cuda}, PyTorch \cite{pytorch}, and driver \cite{nvidia_driver} versions.}
         \label{tab:infra}
     \end{table}
     
\section{Memory Footprint Reduction Ablation Study}
\label{appdix:ablation}

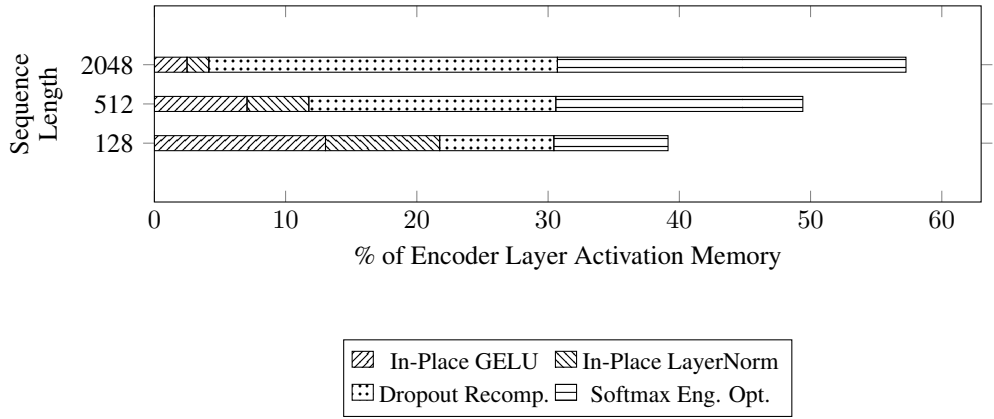
\begin{figure}[!htbp]
\centering
\begin{tikzpicture}
\begin{axis}[
    xbar stacked,
	bar width=.2cm,
	xmin=0,
	width=0.9\columnwidth,
	height=0.3\columnwidth,
    enlarge y limits=0.75,
    legend style={at={(0.5,-0.70)},
      anchor=north,legend columns=2},
    xlabel={\% of Encoder Layer Activation Memory},
    ylabel style={align=center}, ylabel={Sequence\\Length},
    symbolic y coords={128, 512, 2048},
    ytick=data,
    ytick align=outside,
    cycle list={pattern=north east lines, pattern=north west lines, pattern=dots, pattern=horizontal lines},
    ]
\addplot+[xbar] plot coordinates {(13.04,128) (7.06,512) (2.49,2048)};
\addplot+[xbar] plot coordinates {(8.70,128) (4.71,512) (1.66,2048)};
\addplot+[xbar] plot coordinates {(8.70,128) (18.82,512) (26.56,2048)};
\addplot+[xbar] plot coordinates {(8.70,128) (18.82,512) (26.56,2048)};
\legend{\strut In-Place GELU, \strut In-Place LayerNorm,\strut Dropout Recomp., \strut Softmax Eng. Opt.}
\end{axis}
\end{tikzpicture}
\caption{Per layer comparison of Tempo memory footprint reduction across different sequence lengths.}
\label{fig:memory_ablation}
\end{figure}

We calculate the memory footprint reduction contributed by each optimization across different sequence lengths relative to the total memory footprint of each encoder layer, given a $H$ to $A$ ratio of 64 \cite{vaswani2017, devlin2018}. This is shown in Figure \ref{fig:memory_ablation} for the selected configurations. From the figure, it's clear that In-Place GELU and LayerNorm provide the bulk of the memory footprint reduction in the short sequence length regime, while the other two optimizations provide an improvement in the long sequence length case. Note that this is due to the fact that the latter's memory footprint reduction is $\mathcal{O}(S^2)$, while the former's memory footprint reduction goes as $\mathcal{O}(SH)$ where $S$ is the sequence length. This allows \emph{Tempo} to stay robust to model parameters, and provide consistent performance across sequence lengths.


\section{Checklist}
\label{appdix:checklist}

\begin{enumerate}

\item For all authors...
\begin{enumerate}
  \item Do the main claims made in the abstract and introduction accurately reflect the paper's contributions and scope?
    \answerYes{}
  \item Did you describe the limitations of your work?
    \answerYes{}
  \item Did you discuss any potential negative societal impacts of your work?
    \answerNA{This is a system level work which does not have any direct negative impact other than that of the underlying model. We discuss the positive impact in Section 1.}
  \item Have you read the ethics review guidelines and ensured that your paper conforms to them?
    \answerYes{}
\end{enumerate}

\item If you are including theoretical results...
\begin{enumerate}
  \item Did you state the full set of assumptions of all theoretical results?
    \answerYes{}
        \item Did you include complete proofs of all theoretical results?
    \answerYes{See Appendix/}
\end{enumerate}

\item If you ran experiments...
\begin{enumerate}
  \item Did you include the code, data, and instructions needed to reproduce the main experimental results (either in the supplemental material or as a URL)?
    \answerYes{}
  \item Did you specify all the training details (e.g., data splits, hyperparameters, how they were chosen)?
    \answerYes{}
        \item Did you report error bars (e.g., with respect to the random seed after running experiments multiple times)?
    \answerYes{For accuracy experiments, otherwise all others are system level experiments which should not change based on random seed and are expensive to run.}
        \item Did you include the total amount of compute and the type of resources used (e.g., type of GPUs, internal cluster, or cloud provider)?
    \answerYes{See Appendix.}
\end{enumerate}

\item If you are using existing assets (e.g., code, data, models) or curating/releasing new assets...
\begin{enumerate}
  \item If your work uses existing assets, did you cite the creators?
    \answerYes{}
  \item Did you mention the license of the assets?
    \answerNo{}
  \item Did you include any new assets either in the supplemental material or as a URL?
    \answerNo{}
  \item Did you discuss whether and how consent was obtained from people whose data you're using/curating?
    \answerNA{}
  \item Did you discuss whether the data you are using/curating contains personally identifiable information or offensive content?
    \answerNA{}
\end{enumerate}

\item If you used crowdsourcing or conducted research with human subjects...
\begin{enumerate}
  \item Did you include the full text of instructions given to participants and screenshots, if applicable?
    \answerNA{}
  \item Did you describe any potential participant risks, with links to Institutional Review Board (IRB) approvals, if applicable?
    \answerNA{}
  \item Did you include the estimated hourly wage paid to participants and the total amount spent on participant compensation?
    \answerNA{}
\end{enumerate}

\end{enumerate}


\end{document}